%% file: MainPaper.tex
\documentclass[letterpaper]{article} 
\usepackage{aaai24}  
\usepackage{times}  
\usepackage{helvet}  
\usepackage{courier}  
\usepackage[hyphens]{url}  
\usepackage{graphicx} 
\usepackage{natbib}  
\usepackage{caption} 
\usepackage{algorithm}
\usepackage[noend]{algpseudocode}
\usepackage{amsmath}
\usepackage{amsthm}
\usepackage{amssymb}
\usepackage{bm}
\usepackage{subfig}
\usepackage{graphicx}
\usepackage{epsfig}
\usepackage{color}
\usepackage[dvipsnames]{xcolor}
\usepackage{array}
\usepackage{multirow}
\usepackage{makecell}
\usepackage[flushleft]{threeparttable}
\usepackage[textsize=scriptsize]{todonotes}
\usepackage{mathtools}
\DeclareMathOperator*{\argmax}{arg\,max}

\newcommand\hood{\mathcal{S}}

\newcommand{\tb}[1]{$\textbf{#1}$}

\newcommand\etc{\emph{etc.}}

\newtheorem{remark}{Remark}
\newtheorem{theorem}{Theorem}
\newtheorem{lemma}{Lemma}
\newtheorem{definition}{Definition}

\newtheorem{proposition}{Proposition}
\newcommand{\semcaption}[1]{\caption{\small #1}}

\usepackage{newfloat}
\usepackage{listings}
\DeclareCaptionStyle{ruled}{labelfont=normalfont,labelsep=colon,strut=off} 
\floatstyle{ruled}
\newfloat{listing}{tb}{lst}{}
\floatname{listing}{Listing}
\title{Adversarial Purification with the Manifold Hypothesis}
\author{
    Zhaoyuan Yang$^1$, Zhiwei Xu$^2$, Jing Zhang$^2$, Richard Hartley$^2$, Peter Tu$^1$
}
\affiliations{
    \textsuperscript{\rm 1}GE Research~ \textsuperscript{\rm 2}Australian National University\\
\{zhaoyuan.yang,tu\}@ge.com, \{zhiwei.xu,jing.zhang,richard.hartley\}@anu.edu.au
}

\usepackage{bibentry}

\begin{document}

\maketitle

\begin{abstract}
In this work, we formulate a novel framework for adversarial robustness using the manifold hypothesis. This framework provides sufficient conditions for defending against adversarial examples. We develop an adversarial purification method with this framework. Our method combines manifold learning with variational inference to provide adversarial robustness without the need for expensive adversarial training. Experimentally, our approach can provide adversarial robustness even if attackers are aware of the existence of the defense. In addition, our method can also serve as a test-time defense mechanism for variational autoencoders. 
\end{abstract}

\section{Introduction}
\label{sec:intro}

State-of-the-art neural network models are known of being vulnerable to adversarial examples. With small perturbations, adversarial examples can completely change predictions of neural networks \cite{DBLP:journals/corr/SzegedyZSBEGF13}. Defense methods are then designed to produce robust models towards adversarial attacks. Common defense methods for adversarial attacks include adversarial training \cite{madry2018towards}, certified robustness \cite{wong2018provable}, \etc. 
 Recently, adversarial purification has drawn increasing attention \cite{croce2022evaluating},
 which purifies adversarial examples during test time and thus requires fewer training resources. 

Existing adversarial purification methods achieve superior performance when attackers are not aware of the existence of the defense; however, their performance
drops significantly when attackers create defense-aware or adaptive attacks \cite{croce2022evaluating}. Besides, most of them
are empirical with limited
theoretical justifications.
Differently, we adapt ideas from the certified robustness and build an adversarial purification method with a theoretical foundation.

\begin{figure}[t]
	\begin{center}
		\includegraphics[width=0.95\linewidth]{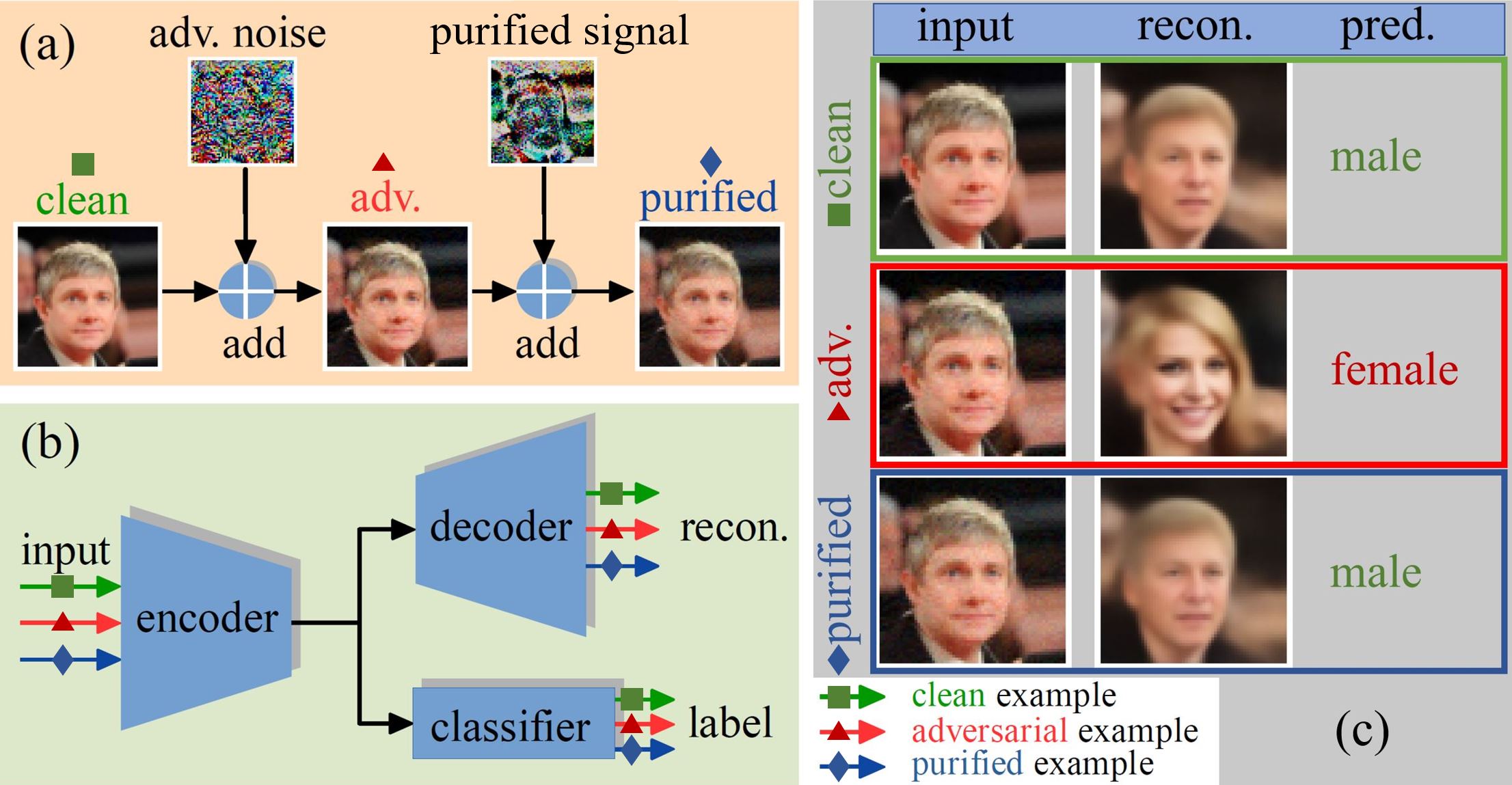}
	\end{center}
	\semcaption{Adversarial purification against adversarial attacks. (a)
 Clean, adversarial (adv.), and purified images.
 (b) Jointly learning of the variational autoencoder and the classifier to achieve semantic consistency. (c) Applying semantic consistency between predictions and reconstructions to defend against attacks. 
 }
	\label{fig:jae_example}
\end{figure}

Specifically, our adversarial purification method is based on the assumption that high-dimensional images lie on low-dimensional manifolds (the manifold hypothesis). Compared with low-dimensional data, high-dimensional data are more vulnerable to adversarial examples \cite{goodfellow2014explaining}. Thus, we transform the adversarial robustness problem from a high-dimensional image domain to a low-dimensional image manifold domain and present a novel adversarial purification method for non-adversarially trained models via manifold learning and variational inference (see Figure~\ref{fig:jae_example} for the pipeline).
With our method, non-adversarially trained models can achieve performance on par with the performance of adversarially trained models. Even if attackers are aware of the existence of the defenses, our approach can still provide adversarial robustness against attacks.

Our method is significant in introducing the manifold hypothesis to the adversarial defense framework. We improve a model's adversarial robustness from a more interpretable low-dimensional image manifold than the complex high-dimensional image space. In the meantime, we provide conditions (in theory) to quantify the robustness of the predictions.
Also, we present an effective adversarial purification approach combining manifold learning and variational inference, which achieves reliable performance on adaptive attacks without adversarial training. We also demonstrate the feasibility of our method to improve the robustness of adversarially trained models. 

\section{Related Work}

\noindent\textbf{Adversarial Training.}
Adversarial training is one of the most effective adversarial defense methods which incorporates adversarial examples into the training set \cite{goodfellow2014explaining,madry2018towards}. Such a method could degrade classification accuracy on clean data \cite{tsipras2018robustness,DBLP:conf/icml/PangLYZY22}. To reduce the degradation in clean classification accuracy, TRADES \cite{zhang2019theoretically} is proposed to balance the trade-off between clean and robust accuracy. Recent works also study the effects of different hyperparameters \cite{pang2020bag,huang2022revisiting} and data augmentation \cite{rebuffi2021fixing,DBLP:conf/iclr/SehwagMHD0CM22,DBLP:conf/nips/000300M20} to reduce robust overfitting and avoid the decrease of model's robust accuracy. Besides the standard adversarial training, many works also study the impact of adversarial training on manifolds \cite{stutz2019disentangling,lin2020dual,zhou2020,patel2020}. Different from these works, we introduce a novel defense without adversarial training.

\noindent\textbf{Adversarial Purification and Test-time Defense.}
As an alternative to adversarial training, adversarial purification aims to shift adversarial examples back to the representations of clean examples. Some efforts perform adversarial purification using GAN-based models~\cite{samangouei2018iclr}, energy-based models \cite{grathwohl2020,yoon2021,hill2021}, autoencoders \cite{hwang2019,DBLP:journals/ijon/YinZZ22,willetts2021,DBLP:conf/iclr/0004YLZLL022,DBLP:conf/ccs/MengC17}, augmentations \cite{DBLP:conf/iccvw/PerezAJRTGA21,shi2020online,mao2021adversarial},  \etc. \citet{song2018pixeldefend} discover that adversarial examples lie in low-probability regions and they use PixelCNN to restore the adversarial examples by shifting them back to high-probability regions. \citet{shi2020online} and \citet{mao2021adversarial} discover that adversarial attacks increase the loss of self-supervised learning and they define reverse vectors to purify the adversarial examples. Prior efforts \cite{athalye2018obfuscated,croce2022evaluating} have shown that methods such as Defense-GAN, PixelDefend (PixelCNN), SOAP (self-supervised), autoencoder-based purification are vulnerable to the Backward Pass Differentiable Approximation (BPDA) attacks \cite{athalye2018obfuscated}. Recently, diffusion-based adversarial purification methods have been studied~\cite{nie2022diffusion,xiao2023densepure} and show adversarial robustness against adaptive attacks such as BPDA. \citet{lee2023robust}, however, observe that the robustness of diffusion-based purification drops significantly when evaluated with the surrogate gradient designed for diffusion models. Similar to adversarial purification, existing test-time defense techniques~\cite{nayak2022dad,huang2023testtime} are also vulnerable to adaptive white-box attacks. In this work, we present a novel defense method combining manifold learning and variational inference which achieves better performance compared with prior works and greater robustness on adaptive white-box attacks. 

\section{Methodology}
In this section, we introduce an adversarial purification method with the manifold hypothesis.
We first define sufficient conditions (in theory) to quantify the robustness of predictions.
Then, we use variational inference to approximate such conditions in implementation and achieve adversarial robustness without adversarial training.

Let $\mathcal{D}_{XY}$ be a set of clean images and their labels where each image-label pair is defined as $(\mathbf{x}, \mathbf{y})$. The manifold hypothesis states that many real-world high-dimensional data $\mathbf{x} \in \mathbb{R}^n$ lies on a low-dimensional manifold $\mathcal{M}$ diffeomorphic to $\mathbb{R}^m$ with $m \ll n$. We define an encoder function $\mathbf{f}:\mathbb{R}^n \to \mathbb{R}^m$ and a decoder function $\mathbf{f}^{\dagger}:\mathbb{R}^m \to \mathbb{R}^n$ to form an autoencoder, where $\mathbf{f}$ maps data point $\mathbf{x} \in \mathbb{R}^n$ to point $\mathbf{f}(\mathbf{x}) \in \mathbb{R}^m$. For $\mathbf{x} \in \mathcal{M}$, $\mathbf{f}^{\dagger}$ and $\mathbf{f}$ are approximate inverses (see Appendix A for notation details).

\subsection{Problem Formulation}

Let $\mathcal{L}=\{1,...,c\}$ be a discrete label set of $c$ classes and $\mathbf{h}: \mathbb{R}^m \to \mathcal{L}$ be a classifier of the latent space. Given an image-label pair $(\mathbf{x}, \mathbf{y}) \in \mathcal{D}_{XY}$, the encoder maps the image $\mathbf{x}$ to a lower-dimensional vector $\mathbf{z} = \mathbf{f}(\mathbf{x}) \in \mathbb{R}^m$ and the functions $\mathbf{f}$ and $\mathbf{h}$ form a classifier of the image space $\mathbf{y}_{\rm pred} = \mathbf{h}(\mathbf{z}) =  (\mathbf{h} \circ \mathbf{f})(\mathbf{x})$. Generally, the classifier predicts labels consistent with the ground truth labels such that $\mathbf{y}_{\rm pred}=\mathbf{y}$. However, during adversarial attacks, the adversary can generate a small adversarial perturbation $\bm \delta_{\rm adv}$ such that $(\mathbf{h} \circ \mathbf{f})(\mathbf{x}) \neq (\mathbf{h} \circ \mathbf{f})(\mathbf{x} + \bm \delta_{\rm adv})$. Thus, our purification framework aims to find a purified signal $\bm \epsilon_{\rm pfy} \in \mathbb{R}^n$ such that $(\mathbf{h} \circ \mathbf{f})(\mathbf{x}) = (\mathbf{h} \circ \mathbf{f})(\mathbf{x} + \bm \delta_{\rm adv} + \bm \epsilon_{\rm pfy})=\mathbf{y}$. However, it is challenging to achieve $\bm \epsilon_{\rm pfy}=-\bm \delta_{\rm adv}$ because $\bm \delta_{\rm adv}$ is unknown. Thus, we aim to seek an alternative approach to estimate the purified signal $\bm \epsilon_{\rm pfy}$ and defend against attacks.

\subsection{Theoretical Foundation for Adversarial Robustness}
\label{sec_theory}
The adversarial perturbation is usually $\ell_p$-bounded where $p \in \{0,2,\infty\}$. We define the $\ell_p$-norm of a vector $\mathbf{a}=[a_1,...,a_n]^\intercal$ as $\|\mathbf{a}\|_p$ and a classifier of the image space as $\mathbf{G}: \mathbb{R}^n \to \mathcal{L}$. We follow \citet{DBLP:conf/nips/BastaniILVNC16,DBLP:conf/icml/LeinoWF21} to define the local robustness.

\begin{definition}
(Locally robust image classifier) Given an image-label pair $(\mathbf{x},\mathbf{y})\in \mathcal{D}_{XY}$, a classifier $\mathbf{G}$ is $(\mathbf{x}, \mathbf{y}, \tau)$-robust with respect to $\ell_p$-norm if for every $\bm \eta \in \mathbb{R}^n$ with $\|\bm \eta\|_p \leq \tau$, $\mathbf{y}=\mathbf{G}(\mathbf{x})=\mathbf{G}(\mathbf{x} + \bm \eta)$.
\end{definition}
Human vision is robust up to a certain perturbation budget. For example, given a clean MNIST image $\mathbf{x}$ with pixel values in [0, 1], if $\|\bm \eta\|_\infty \leq 85/255$, human vision will assign $(\mathbf{x} + \bm \eta)$ and $\mathbf{x}$ to the same class \cite{madry2018towards}. We use $\rho_H(\mathbf{x}, \mathbf{y})$ to represent the maximum perturbation budget for static human vision interpretations given an image-label pair $(\mathbf{x},\mathbf{y})$. Exact $\rho_H(\mathbf{x}, \mathbf{y})$ is often a large value but difficult to estimate. We use it to represent the upper bound of achievable robustness.

\begin{definition}
(Human-level image classifier) For every image-label pair $(\mathbf{x}_i,\mathbf{y}_i)\in \mathcal{D}_{XY}$, if a classifier $\mathbf{G}_R$ is $(\mathbf{x}_i, \mathbf{y}_i, \tau_i)$-robust and $\tau_i\triangleq\rho_H(\mathbf{x}_i, \mathbf{y}_i)$ where $\rho_H(\cdot,\cdot)$ represents the maximum perturbation budget for static human vision interpretations, we define such a classifier as a human-level image classifier. 
\end{definition}
The human-level image classifier $\mathbf{G}_R$ is an ideal classifier that is comparable to human vision. To construct such $\mathbf{G}_R$, we need a robust encoder $\mathbf{f}_R$ of the image space and a robust classifier $\mathbf{h}_R$ on the manifold to form $\mathbf{G}_R=\mathbf{h}_R \circ \mathbf{f}_R$. However, it is a challenge to construct a robust encoder $\mathbf{f}_R$ due to the high-dimensional image space. Therefore, we aim to find an alternative solution to enhance the robustness of $\mathbf{h} \circ \mathbf{f}$ against adversarial attacks by enforcing the semantic consistency between the decoder $\mathbf{f}^{\dagger}$ and the classifier $\mathbf{h}$. Both functions ($\mathbf{f}^{\dagger}$ and $\mathbf{h}$) take inputs from a lower dimensional space (compared with the encoder); thus, they are more reliable~\cite{goodfellow2014explaining}.

We define a semantically consistent classifier on the manifold as $\mathbf{h}_S: \mathbb{R}^m \to \mathcal{L}$, which yields a class prediction $\mathbf{h}_S(\mathbf{z})$ given a latent vector $\mathbf{z} \in \mathbb{R}^m$.
\begin{definition}
(Semantically consistent classifier on the manifold $\mathcal{M}$) A semantically consistent classifier $\mathbf{h}_S$ on the manifold $\mathcal{M}$ satisfies the following condition: for all $\mathbf{z} \in \mathbb{R}^m$, $\mathbf{h}_S(\mathbf{z}) = (\mathbf{G}_R\circ \mathbf{f}^{\dagger})(\mathbf{z})$.
\end{definition}
A classifier (on the manifold) is a semantically consistent classifier if its predictions are consistent with the semantic interpretations of the images reconstructed by the decoder. While this definition uses the human-level image classifier $\mathbf{G}_R$, we can use the Bayesian method to approximate $\mathbf{h}_S$ without using $\mathbf{G}_R$ in experiment. Below, we provide the sufficient conditions of adversarial robustness for $\mathbf{h}_S\circ \mathbf{f}$ given an input $\mathbf{x}$, where the encoder $\mathbf{f}$ is not adversarially robust. 

\begin{proposition}
Let $(\mathbf{x},\mathbf{y})$ be an image-label pair from $\mathcal{D}_{XY}$ and the human-level image classifier $\mathbf{G}_R$ be $(\mathbf{x}, \mathbf{y}, \tau)$-robust. If the encoder $\mathbf{f}$ and the decoder $\mathbf{f}^{\dagger}$ are approximately invertible for the given $\mathbf{x}$ such that the reconstruction error $\|\mathbf{x} - (\mathbf{f}^{\dagger}\circ \mathbf{f})(\mathbf{x})\|_p \triangleq \kappa \leq \tau$ (\textbf{sufficient condition}), then there exists a function $\mathbf{F}:\mathbb{R}^n \to \mathbb{R}^n$ such that $(\mathbf{h}_S\circ \mathbf{f} \circ \mathbf{F})$ is $(\mathbf{x}, \mathbf{y}, \frac{\tau - \kappa}{2})$-robust. {\normalfont(See Appendix B)}
\end{proposition}
The function $\mathbf{F}$ is considered to be the purifier for adversarial attacks. We construct such a function based on reconstruction errors. We assume the sufficient condition holds (bounded reconstruction errors $\kappa$ for clean inputs). Lemma~\ref{lma:lemma_1} states that adversarial attacks on a semantically consistent classifier lead to reconstruction errors larger than $\kappa$ (abnormal reconstructions on adversarial examples). 
\begin{lemma}
\label{lma:lemma_1}
If an adversarial example $\mathbf{x}_{\rm adv}=\mathbf{x} + \bm \delta_{\rm adv}$ with $\|\bm \delta_{\rm adv}\|_p \leq  \frac{\tau - \kappa}{2}$ causes $(\mathbf{h}_S \circ \mathbf{f})(\mathbf{x}_{\rm adv}) \neq \mathbf{G}_R(\mathbf{x}_{\rm adv})$, then $\|\mathbf{x}_{\rm adv} - (\mathbf{f}^{\dagger} \circ \mathbf{f})(\mathbf{x}_{\rm adv})\|_p > \frac{\tau+\kappa}{2} \geq \kappa$. {\normalfont(See Appendix B)}
\end{lemma}
To defend against the attacks, we need to reduce the reconstruction error. Theorem~\ref{thm:theorem_1} states that if a purified sample $\mathbf{x}_{\rm pfy}=\mathbf{x}_{\rm adv} + \bm \epsilon_{\rm pfy}$ has a reconstruction error no larger than $\kappa$, the prediction from $(\mathbf{h}_S \circ \mathbf{f})(\mathbf{x}_{\rm pfy})$ will be the same as the prediction from $\mathbf{G}_R(\mathbf{x})$.
\begin{theorem}
\label{thm:theorem_1}
If a purified signal $\bm{\epsilon}_{\rm pfy} \in \mathbb{R}^n$ with $\|\bm{\epsilon}_{\rm pfy}\|_p \leq  \frac{\tau - \kappa}{2}$ ensures that $\|(\mathbf{x}_{\rm adv} + \bm{\epsilon}_{\rm pfy}) - (\mathbf{f}^{\dagger} \circ \mathbf{f})(\mathbf{x}_{\rm adv} + \bm{\epsilon}_{\rm pfy})\|_p \leq \kappa$, then $(\mathbf{h}_S \circ \mathbf{f})(\mathbf{x}_{\rm adv} + \bm{\epsilon}_{\rm pfy}) = \mathbf{G}_R(\mathbf{x})$.{\normalfont(See Appendix B)}
\end{theorem}
If $\bm \epsilon_{\rm pfy}=-\bm \delta_{\rm adv}$, then $\|\mathbf{x}_{\rm pfy} - (\mathbf{f}^{\dagger} \circ \mathbf{f})(\mathbf{x}_{\rm pfy})\|_p = \kappa$. Thus, feasible regions for $\bm \epsilon_{\rm pfy}$ are non-empty. Let $\mathbf{S}:\mathbb{R}^n \to \mathbb{R}^n$ be a function that takes an input $\mathbf{x}$ and outputs a purified signal $\bm \epsilon_{\rm pfy}=\mathbf{S}(\mathbf{x})$ by minimizing the reconstruction error, then $\mathbf{F}(\mathbf{x})\triangleq\mathbf{x}+\mathbf{S}(\mathbf{x})$ and $\mathbf{h}_S \circ \mathbf{f} \circ \mathbf{F}$ is $(\mathbf{x}, \mathbf{y}, \frac{\tau - \kappa}{2})$-robust.

\begin{remark}
\label{rk:remark_1}
For every perturbation $\bm \delta \in \mathbb{R}^n$ with $\|\bm \delta\|_p \leq \nu$, if $\mathbf{S}(\mathbf{x} + \bm \delta)=-\bm \delta$, then the function $\mathbf{S}:\mathbb{R}^n \to \mathbb{R}^n$ is locally Lipschitz continuous on $\mathcal{B}_\nu \triangleq \{\hat{\mathbf{x}} \in \mathbb{R}^n \mid \|\hat{\mathbf{x}} - \mathbf{x}\|_p < \nu\}$ with a Lipschitz constant of 1.{\normalfont(See Appendix B)}
\end{remark}

\noindent\textbf{Insights From the Theory.} Our framework transforms a high-dimensional adversarial robustness problem into a low-dimensional semantic consistency problem. Since we only provide the sufficient conditions for adversarial robustness, dissatisfaction with the conditions is not necessary to be adversarially vulnerable. Our conditions indicate that higher reconstruction quality could lead to stronger robustness. Meanwhile, our method can certify robustness up to $\frac{\tau}{2}$ (reconstruction error $\kappa=0$) when a human-level image classifier can certify robustness up to $\tau$. The insight is that adding a purified signal $\bm \epsilon_{\rm pfy}$ on top of an adversarial example $\mathbf{x}_{\rm adv}$ could change the image semantic, see Figure~\ref{fig:cifar_AE_VAE}(c). In our formulation, we use $\tau$ to quantify the local robustness of the human-level image classifier $\mathbf{G}_R$ given an image-label pair $(\mathbf{x},\mathbf{y})$. Our objective is not to estimate the $\tau$ value but to use it to represent the upper bound of the achievable local robustness. Since $\mathbf{G}_R$ can be considered as a human vision model, the $\tau$ value is often a large number. Our framework is based on the triangle inequality of the $\ell_p$-metrics; thus, it can be extended to other distance metrics.

\begin{figure}[t]
	\begin{center}
		\includegraphics[width=\linewidth]{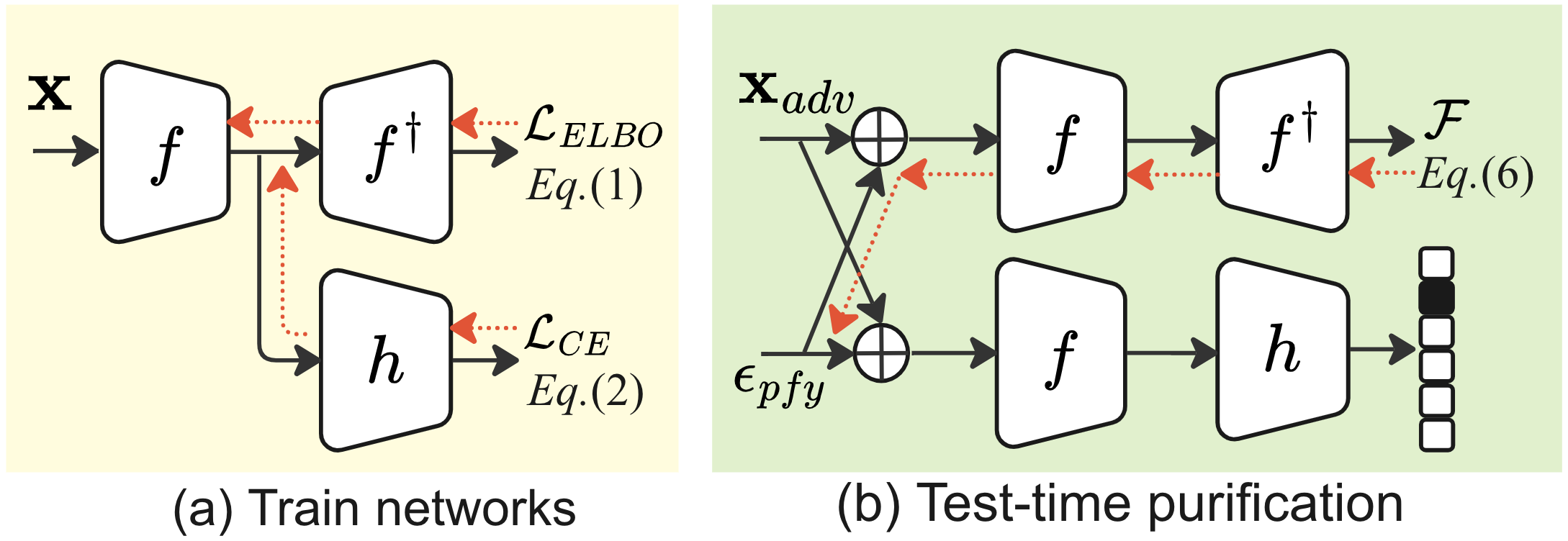}
	\end{center}
	\semcaption{Two-stage pipeline: (a) jointly training for semantic consistency between the decoder and the classifier and (b) iterative updates of $\epsilon_{\text{pfy}}$ to purify $\mathbf{x}_{\text{adv}}$ in inference.}
	\label{fig:flow_chart}
\end{figure}

\noindent\textbf{Relaxation.} Our framework requires semantic consistency between the classifier on the manifold and the decoder on the manifold. Despite that the classifiers and decoders (on the manifold) have a low input dimension, it is still difficult to achieve high semantic consistency between them. Meanwhile, the human-level image classifier $\mathbf{G}_R$ is not available. Thus, we assume that predictions and reconstructions from high data density regions of $p(\mathbf{z}|\mathbf{x})$ are more likely to be semantically consistent \cite{DBLP:conf/cvpr/Zhou22}. In the following sections, we introduce a practical implementation of adversarial purification based on our theoretical foundation. The implementation includes two stages: (1) enforce semantic consistency during training and (2) test-time purification of adversarial examples, see Figure~\ref{fig:flow_chart}.

\subsection{Semantic Consistency with the ELBO}

Exact inference of $p(\mathbf{z}|\mathbf{x})$ is often intractable, we, therefore, use variational inference to approximate the posterior $p(\mathbf{z}|\mathbf{x})$ with a different distribution $q(\mathbf{z}|\mathbf{x})$. We define two parameters $\theta$ and $\phi$ which parameterize the distributions $p_{\theta}(\mathbf{x} | \mathbf{z})$ and $q_{\phi}(\mathbf{z} | \mathbf{x})$. When the evidence lower bound (ELBO) is maximized, $q_{\phi}(\mathbf{z} | \mathbf{x})$ is considered to be a reasonable approximation of $p(\mathbf{z}|\mathbf{x})$. To enforce the semantic consistency between the classifier and the decoder, we force the latent vector $\mathbf{z}$ inferred from the $q_{\phi}(\mathbf{z} | \mathbf{x})$ to contain the class label information of the input $\mathbf{x}$. We define a one-hot label vector as $\mathbf{y} = [y_1, y_2 ..., y_c]^\intercal$, where $c$ is the number of classes and $y_i = 1$ if the image label is $i$ otherwise it is zero. A classification head parametrized by $\bm{\psi}$ is represented as $\mathbf{h}_{\bm{\psi}}(\mathbf{z}) = [h_1(\mathbf{z}), h_2(\mathbf{z}) ..., h_c(\mathbf{z})]^\intercal$ and the cross-entropy classification loss is $-\mathbb{E}_{\mathbf{z} \sim q_{\bm{\phi}}(\mathbf{z} | \mathbf{x} )}[\mathbf{y}^\intercal \log \mathbf{h}_{\bm{\psi}}(\mathbf{z})]$ where $\log(\cdot)$ is an element-wise function for a vector. We assume the classification loss is no greater than a threshold $\omega$ and the training objective can then be written as 
\begin{align}
&\max_{\bm{\theta} , \bm{\phi}} \underbrace{\mathbb{E}_{\mathbf{z} \sim q_{\bm{\phi}}(\mathbf{z} | \mathbf{x} )}{\left[\log p_{\bm{\theta}}(\mathbf{x} | \mathbf{z})\right]}   - D_{KL}[q_{\bm{\phi}}(\mathbf{z} | \mathbf{x} ) \| p(\mathbf{z})]}_{\text{ELBO (lower bound of} \log p_{\bm{\theta}}(\mathbf{x}) \text{)}}\  \\
&\text{ s.t. } -\mathbb{E}_{\mathbf{z} \sim q_{\bm{\phi}}(\mathbf{z} | \mathbf{x} )}[\mathbf{y}^\intercal \log \mathbf{h}_{\bm{\psi}}(\mathbf{z})] \leq \omega.
\label{eq:joint_elbo_cls}
\end{align}

We use the Lagrange multiplier with KKT conditions to optimize this objective as
\begin{equation}
\max_{\bm{\theta} ,\bm{\phi}, \bm{\psi}} \text{ELBO} + \lambda \mathbb{E}_{\mathbf{z} \sim q_{\bm{\phi}}(\mathbf{z} | \mathbf{x} )}[\mathbf{y}^\intercal \log \mathbf{h}_{\bm{\psi}}(\mathbf{z})],
\label{eq:joint_elbo_cls_kkt}
\end{equation}
where $\lambda$ is a trade-off term to balance the two loss terms. 

We follow \citet{DBLP:journals/corr/KingmaW13} to define the prior $p(\mathbf{z})=\mathcal{N}(0,I)$ and the posterior (encoder) $q_{\phi}(\mathbf{z}|\mathbf{x})$ by using a normal distribution with diagonal covariance. Given an input vector $\mathbf{x}$, an encoder model parameterized by $\phi$ is used to model the posterior distribution $q_{\phi}(\mathbf{z}|\mathbf{x})=\mathcal{N}(\bm{\mu}_{\bm{\phi}}(\mathbf{x}), \text{diag}(\bm{\sigma}^2_{\bm{\phi}}(\mathbf{x})))$. The model predicts the mean vector $\bm{\mu}_{\bm{\phi}}(\mathbf{x}) = [\mu_1(\mathbf{x}), \mu_2(\mathbf{x})...,\mu_m(\mathbf{x})]^\intercal$ and the diagonal covariance $\bm{\sigma}_{\bm{\phi}}^2(\mathbf{x}) = [\sigma_1^2(\mathbf{x}), \sigma_2^2(\mathbf{x})...,\sigma_m^2(\mathbf{x})]^\intercal$. We define $p_{\bm{\theta}}(\mathbf{x} | \mathbf{z}) = (1/\beta)\text{exp}(-(1/\gamma)\|\mathbf{x} - f_{\bm{\theta}}^{\dagger}(\mathbf{z})\|^2_2)$, where $\gamma$ controls the variance, $\beta$ is a normalization constant, and $f_{\bm{\theta}}^{\dagger}$ is a decoder parametrized by $\bm{\theta}$, which maps data from the latent space to the image space. The illustrated network optimizing over Eq.~\eqref{eq:joint_elbo_cls_kkt} is provided in Figure~\ref{fig:flow_chart}(a). 

\subsection{Adversarial Attack and Purification}
We mainly focus on the white-box attacks here. During inference time, the attackers can create adversarial perturbations on the classification head. Given a clean image $\mathbf{x}$, an adversarial example can be crafted as $\mathbf{x}_{\rm adv} = \mathbf{x} + \bm \delta_{\rm adv}$ with
\begin{equation}
\label{eq:atk_jointly_obj}
\bm{\delta}_{\rm adv} = \argmax_{\bm{\delta} \in \mathcal{C}_{\rm adv}} -\mathbb{E}_{\mathbf{z} \sim q_{\bm{\phi}}(\mathbf{z} | \mathbf{x} + \bm{\delta})}[\mathbf{y}^\intercal \log \mathbf{h}_{\bm{\psi}}(\mathbf{z})],
\end{equation}
where $\mathcal{C}_{\rm adv} \triangleq\{\bm{\delta} \in \mathbb{R}^n \mid \mathbf{x} + \bm{\delta} \in [0,1]^n \text{ and } \|\bm{\delta}\|_{p} \leq \delta_{\rm th} \}$ is the feasible set for $\delta_{\rm th}$-bounded perturbations. 

To avoid changing the semantic interpretation of the image, we need to estimate the purified signal with a $\ell_p$-norm no greater than a threshold value $\epsilon_{\rm th}$. The purified signal also needs to project the sample to a high-density region of the manifold with a small reconstruction loss. The ELBO contains $\ell_2$ reconstruction loss (relaxation from $\ell_p$-metric), and it can be applied to maximize the posterior $q_{\bm{\phi}}(\mathbf{z} | \mathbf{x})$. Therefore, we present a defense method that optimizes the ELBO during the test time to degrade the effects of the attacks. Given an adversarial example $\mathbf{x}_{\rm adv}$, a purified sample can be obtained by $\mathbf{x}_{\rm pfy} = \mathbf{x}_{\rm adv} + \bm{\epsilon}_{\rm pfy}$ with
\begin{equation}
\label{eq:purify_jointly_obj}
\begin{aligned}
\bm{\epsilon}_{\rm pfy} = \argmax_{\bm{\epsilon} \in \mathcal{C}_{\rm pfy}} \mathbb{E}_{\mathbf{z} \sim \hat{q}_{\bm{\phi}}(\mathbf{z})}{\left[\log \hat{p}_{\bm{\theta}}(\bm{\epsilon})\right]} - D_{\rm KL}[\hat{q}_{\bm{\phi}}(\bm{\epsilon}) \| p(\mathbf{z})],
\end{aligned}
\end{equation}
where $\hat{p}_{\bm{\theta}}(\bm{\epsilon}) \triangleq p_{\bm{\theta}}(\mathbf{x}_{\rm adv} + \bm{\epsilon}| \mathbf{z})$, $\hat{q}_{\bm{\phi}}(\bm{\epsilon})\triangleq q_{\bm{\phi}}(\mathbf{z} | \mathbf{x}_{\rm adv} + \bm{\epsilon})$ and $\mathcal{C}_{\rm pfy}\triangleq\{\bm{\epsilon} \in \mathbb{R}^n \mid \mathbf{x}_{\rm adv} + \bm{\epsilon} \in [0,1]^n \text{ and } \|\bm{\epsilon}\|_{p} \leq \epsilon_{\rm th} \}$ which is a feasible set for purification. Compared with training a model to produce the purified signal $\bm \epsilon_{\rm pfy}=\mathbf{S}(\mathbf{x})$, the test-time optimization of the ELBO is more efficient.

We focus on $\ell_{\infty}$-bounded purified vectors while our method is effective for both $\ell_2$ and $\ell_{\infty}$ attacks in our experiments. We define $\alpha$ as the learning rate and $\text{Proj}_{\hood}$ as the projection operator which projects a data point back to its feasible region when it is out of the region. We use a clipping function as the projection operator to ensure $\|\mathbf{x}_{\rm pfy} - \mathbf{x}\|_{\infty} = \|\bm{\epsilon}_{\rm pfy}\|_{\infty} \leq \epsilon_{\rm th}$ and $\mathbf{x}_{\rm pfy} = \mathbf{x} + \bm{\epsilon}_{\rm pfy} \in [0,1]^n$, where $\mathbf{x}$ is an  (adversarial) image and $\epsilon_{\rm th}$ is the budget for purification. We then define $\mathcal{F}(\mathbf{x}; \bm{\theta}, \bm{\phi})$ as
\begin{equation}
 \mathbb{E}_{\mathbf{z} \sim q_{\bm{\phi}}(\mathbf{z} | \mathbf{x} )}{\left[\log p_{\bm{\theta}}(\mathbf{x} | \mathbf{z})\right]} - D_{\rm KL}[q_{\bm{\phi}}(\mathbf{z} | \mathbf{x} ) \| p(\mathbf{z})]
\end{equation}
and iterative purification given adversarial example $\mathbf{x}_{\rm adv}$ as
\begin{equation}
\label{eq:iterative_purify}
\bm{\epsilon}^{t+1} = \text{Proj}_{\hood} \left( \bm{\epsilon}^t +
\alpha\operatorname{sgn}(\nabla_{\bm{\epsilon}^t} \mathcal{F}(\mathbf{x}_{\rm adv} + \bm{\epsilon}^t; \bm{\theta}, \bm{\phi}))\right),
\end{equation}
where the element-wise sign function $\text{sgn}(x)=x/|x|$ if $x$ is non-zero otherwise it is zero. A detailed procedure is provided in Figure~\ref{fig:flow_chart}(b) and Algorithm~\ref{alg:ttd}.

\renewcommand{\algorithmicrequire}{\textbf{Input:}}
\renewcommand{\algorithmicensure}{\textbf{Output:}}
\begin{algorithm}
\caption{Test-time Purification}\label{alg:ttd}
\begin{algorithmic}[1]
\Require $\mathbf{x}_{\rm adv}$: input (adv) data; $\alpha$: learning rate; $T$: number of purification iterations; $\epsilon_{\rm th}$: purification budget.
\Ensure $\mathbf{x}_{\rm pfy}$: purified data; $w$: purification score.
\Procedure{Purify}{$\mathbf{x}, \alpha, T, \epsilon_{\rm th}$} 
\State $\bm \epsilon_{\rm pfy} \sim \mathcal{U}_{[-\epsilon_{\rm th}, \epsilon_{\rm th}]}$ \Comment{random initialization}
\For{t = 1, 2, ..., T}
\State $\bm \epsilon_{\rm pfy} \gets \bm \epsilon_{\rm pfy} + \alpha \cdot \text{sign}(\nabla_{\bm \epsilon_{\rm pfy}}\mathcal{F}(\mathbf{x}_{\rm adv} + \bm \epsilon_{\rm pfy}))$
\State $\bm \epsilon_{\rm pfy} \gets \min(\max(\bm \epsilon_{\rm pfy}, -\epsilon_{\rm th}), \epsilon_{\rm th})$
\State $\bm \epsilon_{\rm pfy} \gets \min(\max(\mathbf{x}_{\rm adv} + \bm \epsilon_{\rm pfy}, 0), 1)-\mathbf{x}_{\rm adv}$
\EndFor\label{euclidendwhile}
\State $\mathbf{x}_{\rm pfy} \gets \mathbf{x}_{\rm adv} + \bm \epsilon_{\rm pfy}$ \Comment{purified data}
\State $w \gets \mathcal{F}(\mathbf{x}_{\rm pfy})$ \Comment{purification score}
\State \textbf{return} $\mathbf{x}_{\rm pfy}, w$ 
\EndProcedure
\end{algorithmic}
\end{algorithm}

\begin{figure}[!t]
\centering
\subfloat[Trajectory: clean-attack-purify]{\label{fig:mnist_reconst}\includegraphics[width=0.263\textwidth]{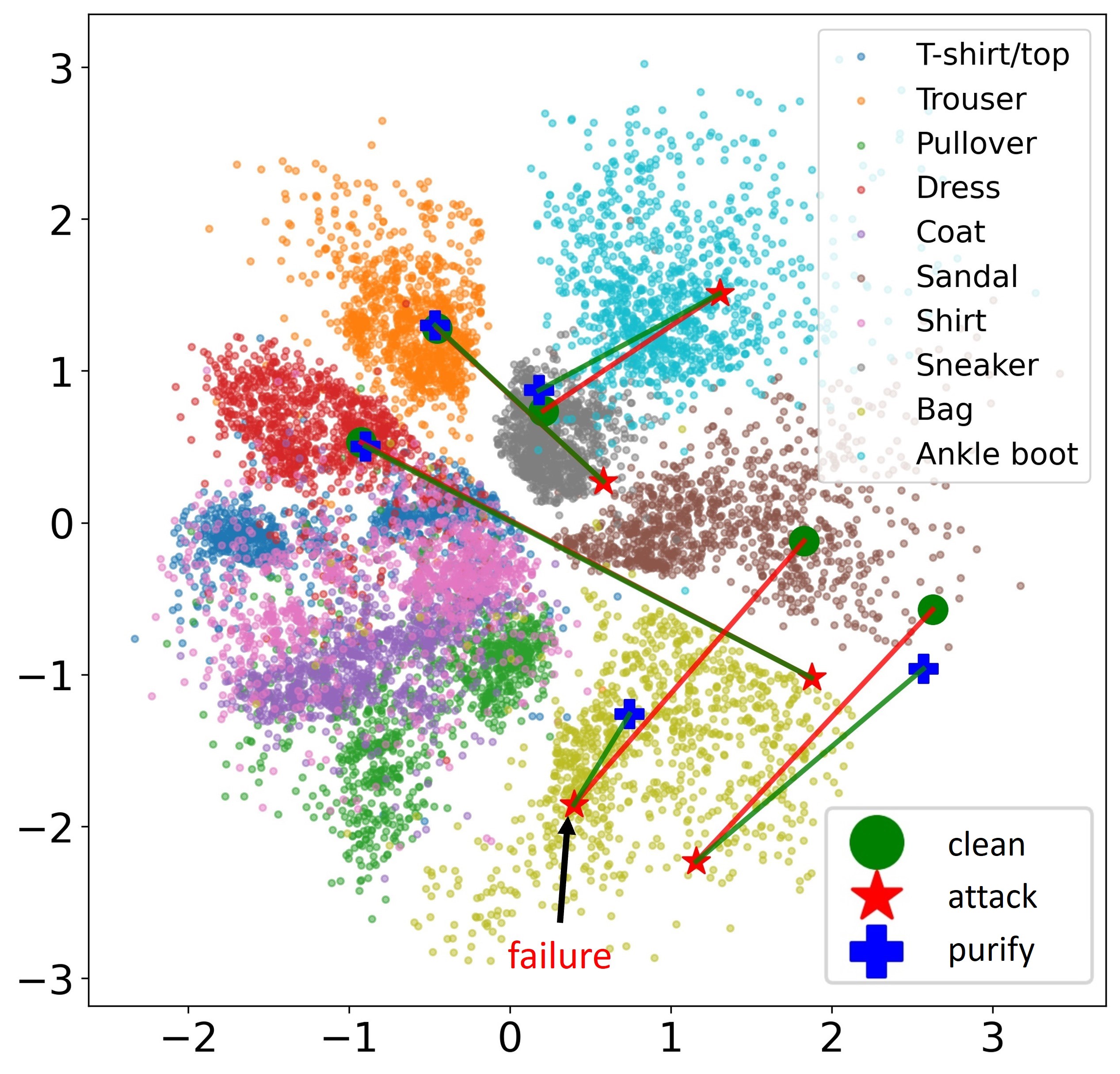}}
\subfloat[3 types of image pairs]{\label{fig:mnist_elbo}\includegraphics[width=0.192\textwidth]{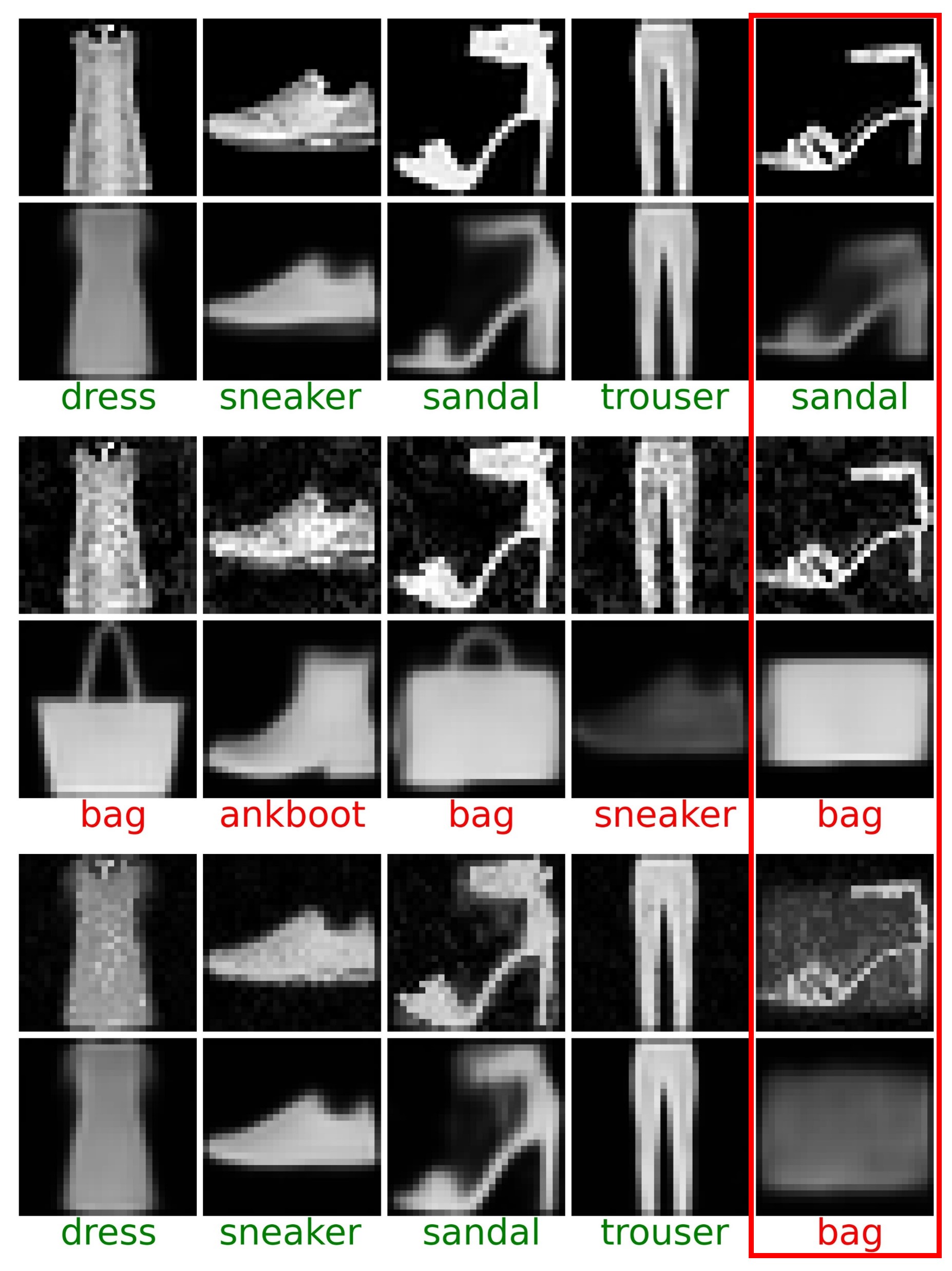}}
\semcaption{(a) 2D trajectories of examples on Fashion-MNIST.
(b) Input and reconstructed image pairs on clean (top), adversarial (middle), and purified (bottom) examples with
correct (green), wrong (red) label predictions and failed cases (in red box).
}
\label{fig:rev_traj}
\end{figure}

The test-time optimization of the ELBO projects adversarial examples back to their feasible regions with a high posterior $q(\mathbf{z}|\mathbf{x})$ (regions where decoders and classifiers have strong semantic consistency) and a small reconstruction error (defend against adversarial attacks). To better demonstrate the process, we build a classification model in a 2-dimensional latent space (Figure~\ref{fig:rev_traj} (a)) on Fashion-MNIST and show examples of clean, attack, and purified trajectories in Figure~\ref{fig:rev_traj}. Correspondingly, adversarial attacks are likely to push latent vectors to abnormal regions which cause abnormal reconstructions (Lemma \ref{lma:lemma_1}). Through the test-time optimization over the ELBO, the latent vectors can be brought back to their original regions (Theorem \ref{thm:theorem_1}).

If attackers are aware of the existence of purification, it is possible to take advantage of this knowledge to create adaptive attacks. A straightforward formulation is to perform the multi-objective attacks with a trade-off term $\lambda_a$ to balance the classification loss $\mathcal{H}(\mathbf{x},\mathbf{y}; \bm{\theta}, \bm{\psi})$ and the purification objective $\mathcal{F}(\mathbf{x}; \bm{\theta}, \bm{\phi})$ \cite{mao2021adversarial,shi2020online}. The adversarial perturbation of the multi-objective attacks is
\begin{equation}
\label{eq:multi_atk_obj_delta_adv}
\bm{\delta}_{\rm adv} = \argmax_{\bm{\delta} \in \mathcal{C}_{\rm adv}} \mathcal{H}(\mathbf{x}+\bm{\delta},\mathbf{y}; \bm{\theta}, \bm{\psi}) + \lambda_a \mathcal{F}(\mathbf{x} + \bm{\delta}; \bm{\theta}, \bm{\phi}).
\end{equation}

Another popular adaptive attack is the BPDA attack \cite{athalye2018obfuscated}. Consider a purification process as $\mathbf{x}_{\rm pfy} = \mathbf{F}(\mathbf{x})$ and a classifier as $\mathbf{G}(\mathbf{x}_{\rm pfy}) = (\mathbf{G} \circ \mathbf{F})(\mathbf{x})$. The BPDA attack uses an approximation $\nabla_\mathbf{\hat{x}} \mathbf{F}(\mathbf{\hat{x}}) \approx \mathbf{I}$ (identity) to estimate the gradient as $\nabla_\mathbf{\hat{x}} \mathbf{G}(\mathbf{F}(\mathbf{\hat{x}}))|_{\mathbf{\hat{x}}={\mathbf{x}}} \approx \nabla_\mathbf{\hat{x}} \mathbf{G}(\mathbf{\hat{x}})|_{\mathbf{\hat{x}}=\mathbf{F}({\mathbf{x}})}$. Many adversarial purification methods are vulnerable to the BPDA attack \cite{athalye2018obfuscated,croce2022evaluating}. We show that even if attackers are aware of the defense, our method can still achieve effective robustness to this white-box adaptive attacks. 

\section{Experiments}
We first evaluate our method on MNIST \cite{lecun2010mnist}, Fashion-MNIST \cite{xiao2017/online}, SVHN \cite{netzer2011reading}, and CIFAR-10 \cite{Krizhevsky09learningmultiple}, followed by CIFAR-100 \cite{Krizhevsky09learningmultiple} and CelebA (64$\times$64 and 128$\times$128) for gender classification \cite{liu2015faceattributes}. See Appendix C for the dataset details.

\noindent\textbf{Model Architectures and Hyperparameters.}
We use three types of models (encoders) in our experiments: (1) tiny ResNet with standard training (for ablation study), (2) ResNet-50 with standard training \cite{he2016deep} (for comparison with standard benchmark), and (3) PreActResNet-18 with adversarial training \cite{rebuffi2021fixing} (to study the impact on adversarially trained models). We use several residual blocks to construct decoders and use linear layers for classification. We empirically set the weight of the classification loss ($\lambda$ in Eq.~\eqref{eq:joint_elbo_cls_kkt}) to 8. See Appendix D for details.

\noindent\textbf{Adversarial Attacks.}
We evaluate our method on standard adversarial attacks and adaptive attacks (multi-objective and BPDA). All attacks are untargeted.
For standard adversarial attacks (Eq.~\eqref{eq:atk_jointly_obj}), we use Foolbox \cite{rauber2017foolbox} to generate the FGSM ($\ell_\infty$) attacks \cite{goodfellow2014explaining}, the PGD ($\ell_\infty$) attacks \cite{madry2018towards}, and the C\&W ($\ell_2$) attacks \cite{carlini2017towards}. We use \citet{croce2020reliable} for the AutoAttack ($\ell_\infty$, $\ell_2$). For the adaptive attacks, we use Torchattacks~\cite{kim2020torchattacks} for the BPDA-PGD/APGD ($\ell_\infty$) attacks \cite{athalye2018obfuscated}, and standard PGD ($\ell_\infty$) for the multi-objective attacks.

For MNIST and Fashion-MNIST, we set the $\ell_\infty$ attack budget $\delta_{\rm th}$ to $50/255$ and the $\ell_2$ attack budget to 3. In Figure~\ref{fig:cifar_AE_VAE}(c), we show that the purification on a larger attack budget could change image semantics. The PGD ($\ell_\infty$) attack is conducted in 200 iterations with step size $2/255$. For the BPDA attack, we use 100 iterations with step size $2/255$.

For SVHN, CIFAR-10/100, and CelebA, we set the $\ell_\infty$ attack budget $\delta_{\rm th}$ to $8/255$ and the $\ell_2$ attack budget to 0.5. We run 100 iterations with step size $2/255$ for PGD ($\ell_\infty$) and 50 iterations with step size $2/255$ for the BPDA attack. 

Meanwhile, we also evaluate our ResNet50 (CIFAR-10) model on the RayS attack~\cite{DBLP:conf/kdd/ChenG20} for model robustness under the blackbox setting.

\begin{figure}[!t]
	\begin{center}
		\includegraphics[width=0.90\linewidth]{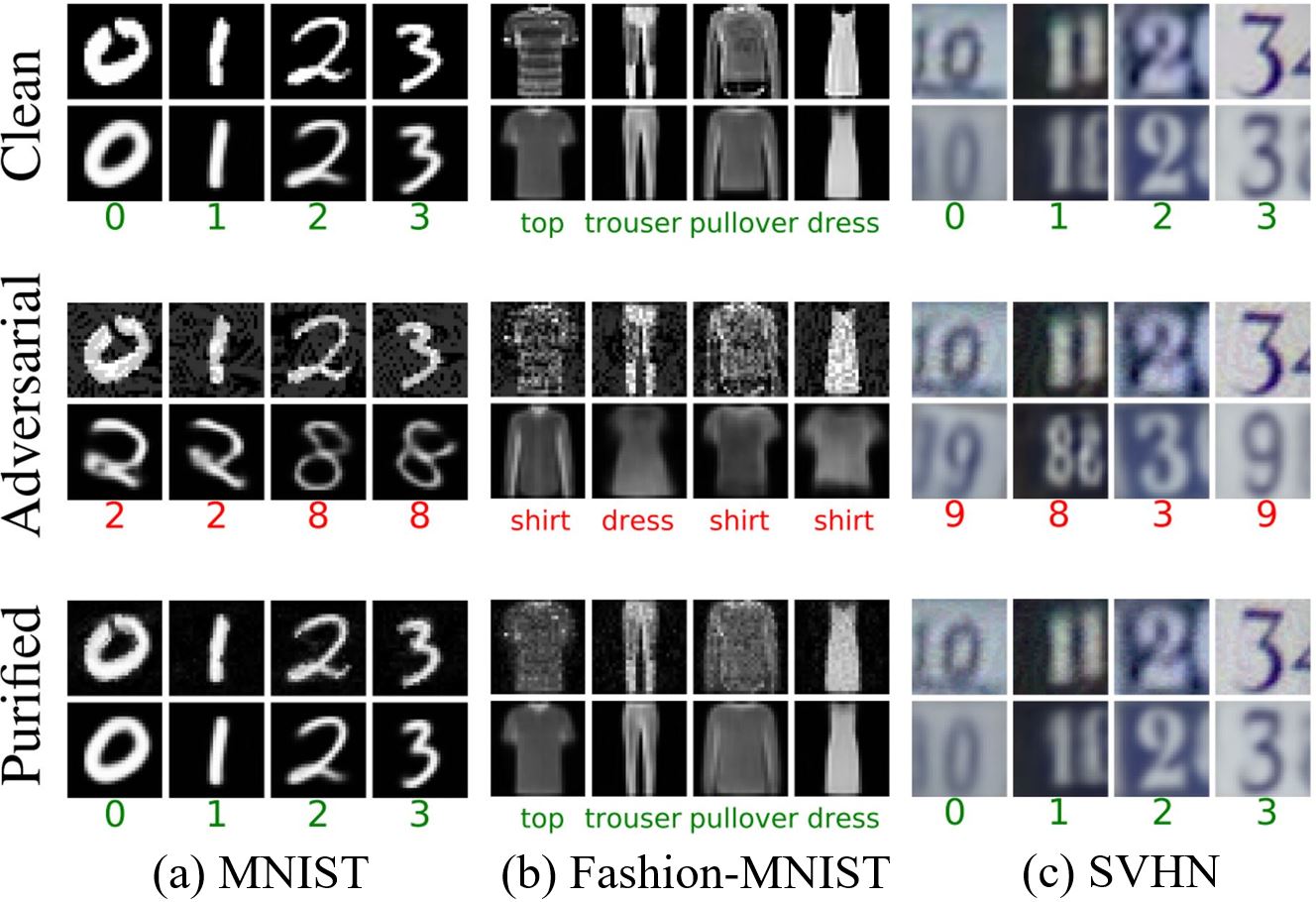}
	\end{center}
	\semcaption{Examples using the VAE-Classifier models on clean, adversarial, and purified images with correct (green) and wrong (red) label predictions, similar to Figure~\ref{fig:rev_traj}(b).
 }
	\label{fig:VAE_atk}
\end{figure}

\begin{figure}[t]
	\begin{center}
		\includegraphics[width=\linewidth]{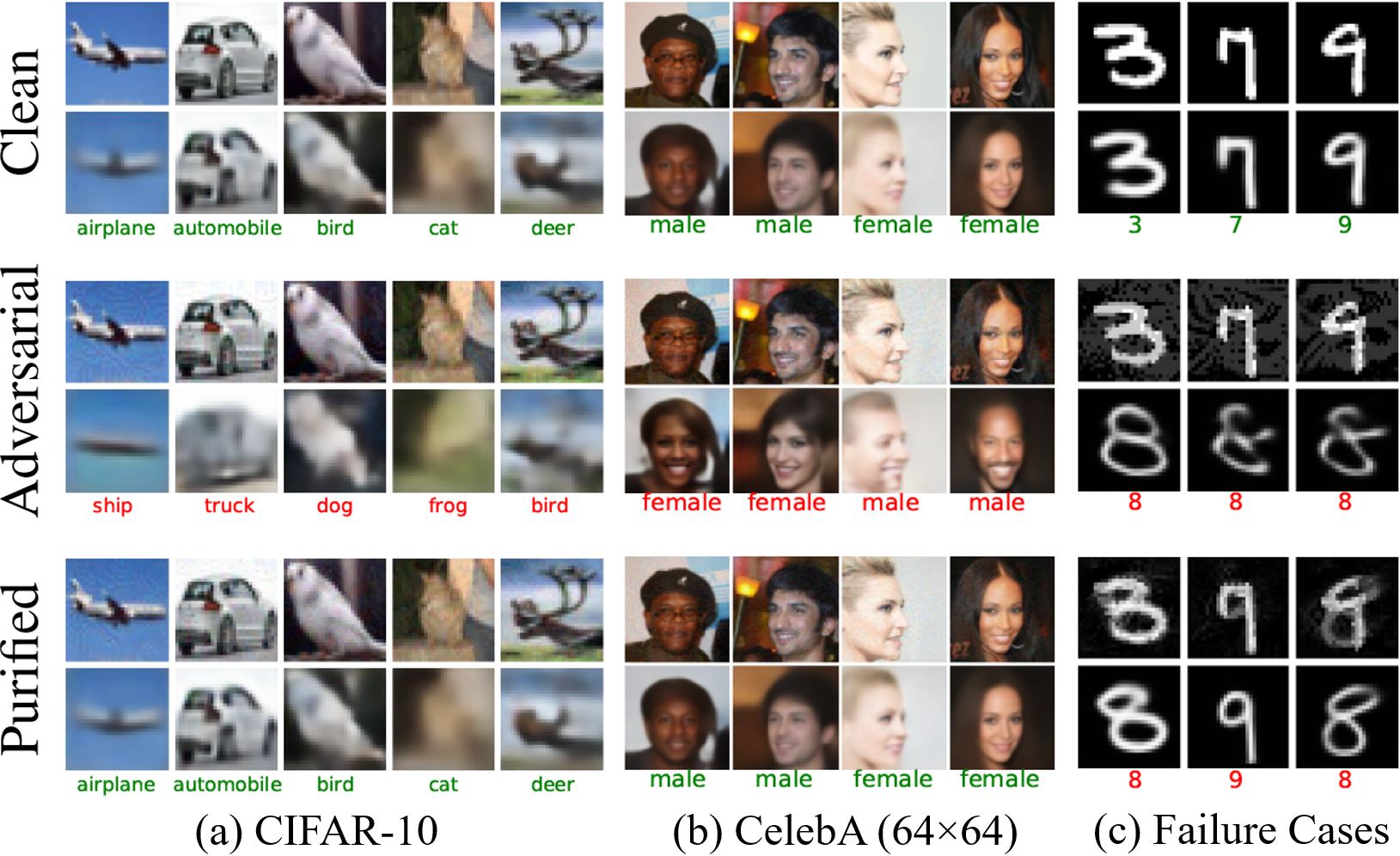}
	\end{center}
	\semcaption{
 Examples of more datasets with failed cases on MNIST in (c) in addition to Figure~\ref{fig:VAE_atk}.
 }
	\label{fig:cifar_AE_VAE}
\end{figure}

\noindent\textbf{Test-time Purification.}
Key hyperparameters and experimental details are provided below, and only the $\ell_\infty$-bounded purification is considered in this work.
We initialize the purified signal $\bm{\epsilon}_{\rm pfy}$ by sampling from an uniform distribution  $\mathcal{U}_{[-\epsilon_{\rm th}, \epsilon_{\rm th}]}$ where $\epsilon_{\rm th}$ is the purification budget. We run purification 16 times in parallel with different initializations and select the best purification score measured by the reconstruction loss or the ELBO. Step size $\alpha$ is alternated between $\{1/255$, $2/255\}$ for each run. See Appendix B for details.

For MNIST and Fashion-MNIST, we set the $\ell_\infty$-purification budget $\epsilon_{\rm th}$ to $50/255$ with 96 purification iterations.
For SVHN, CIFAR-10/100, and CelebA, we set the $\ell_\infty$-purification budget $\epsilon_{\rm th}$ to $8/255$ with 32 iterations. 

Despite the aforementioned hyperparameters, we observe from the experiments that our approach also works on other hyperparameter settings as shown in Figure~\ref{fig:ablation_study}.

\noindent\textbf{Baselines.} We compare our VAE-Classifier (VAE-CLF) with the standard autoencoders, denoted by Standard-AE-Classifier (ST-AE-CLF), by replacing the ELBO with reconstruction loss. One should note that the classifiers of the Standard-AE-Classifier may not have consistency with their decoders. See Appendix D for details.

\noindent\textbf{Objectives of Test-time Purification.} The autoencoders are trained with the reconstruction loss and the ELBO respectively. The Standard-AE-Classifier can only minimize the reconstruction loss during inference while the VAE-Classifier can optimize both the reconstruction loss and the ELBO during inference. We use ``TTP (REC)" to represent the test-time optimization on the reconstruction loss and ``TTP (ELBO)" for the test-time optimization on the ELBO.

\noindent\textbf{Inference Time.} We evaluate our method on an NVIDIA Tesla P100 GPU in PyTorch. For ResNet-50 with a batch size of 256 on CIFAR-10, the average run time per batch for 32 purification steps is 17.65s. Following the standard in \cite{croce2022evaluating}, our method is roughly 102$\times$. To reduce the inference time, one can adapt APGD (adaptative learning rate and momentum) during purification. Parallelization is also promising by running multiple purification processes with different initial values (with fewer purification steps and early stop) to reduce the search time.

\subsection{Experiment Results}

\noindent\textbf{Standard Adversarial Attacks.} For the standard adversarial attacks, only the classification heads are attacked. We observe that, for MNIST, Fashion-MNIST, and SVHN, the standard adversarial attacks of the VAE-Classifier create abnormal reconstructed images while this is not applied to the Standard-AE-Classifier.
It indicates that the classifiers and the decoders of the VAE-Classifier are strongly consistent. Figure~\ref{fig:VAE_atk} shows various sample predictions and reconstructions on clean, attack, and purified images from the VAE-Classifier. For clean images, the VAE-Classifier models achieve qualified reconstruction and predictions. For adversarial examples, the VAE-Classifier models generate abnormal reconstructions which are correlated with abnormal predictions from the classifiers (implied by Lemma \ref{lma:lemma_1}). In other words, if the prediction of an adversarial example is 2, the digit on the reconstructed image may look like 2 as well. If we can estimate the purified vectors by minimizing the errors between inputs and reconstructions, the attacks could be defended (implied by Theorem \ref{thm:theorem_1}). In our experiments, the predictions and reconstructions of adversarial examples become normal after using the test-time optimization over the ELBO.

\begin{figure}[!t]
\centering
\subfloat[MNIST]{\label{fig:img1_mnist}\includegraphics[width=0.22\textwidth]{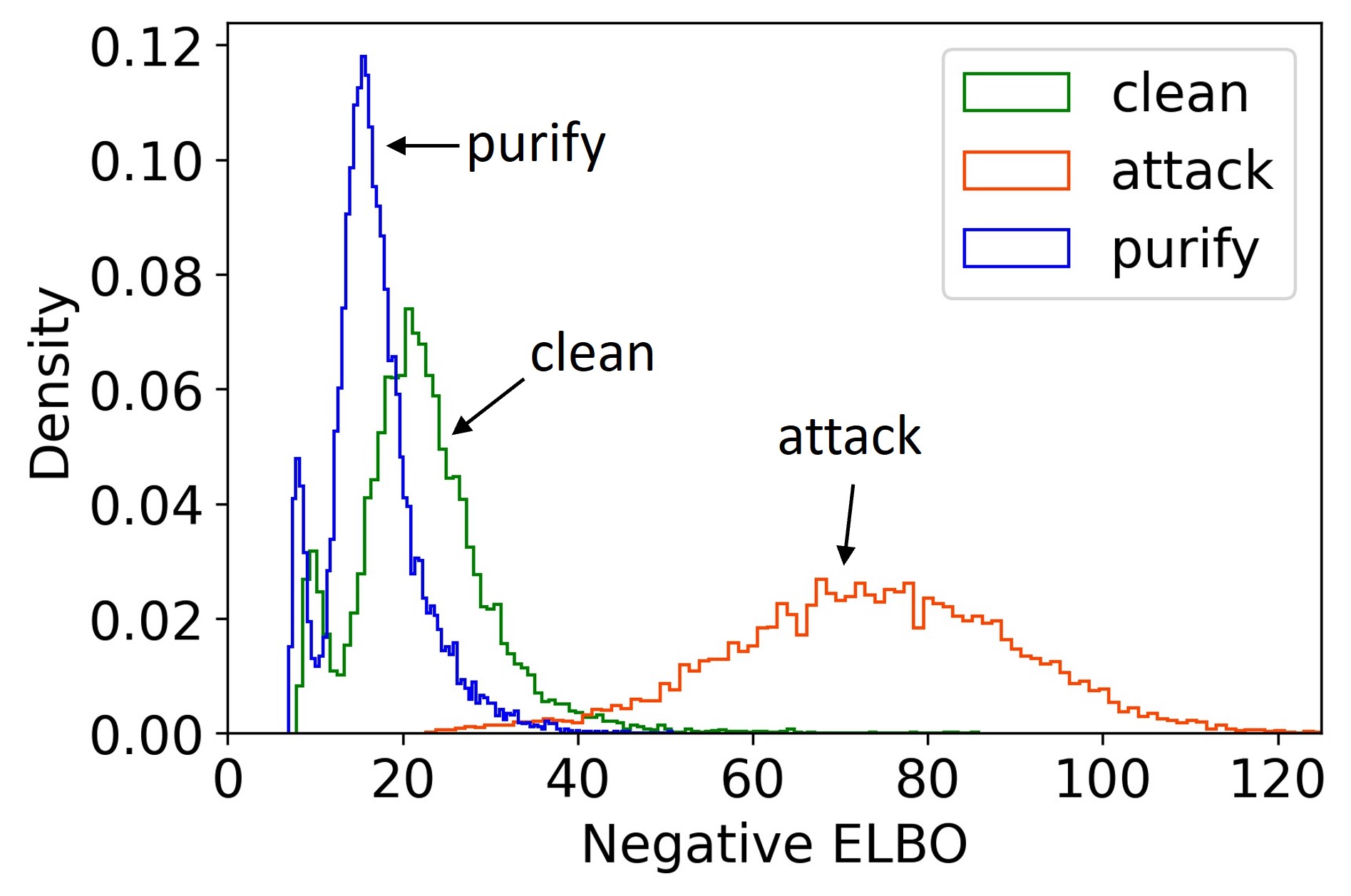}}
\subfloat[CIFAR-10]{\label{fig:img4_cifar10}\includegraphics[width=0.22\textwidth]{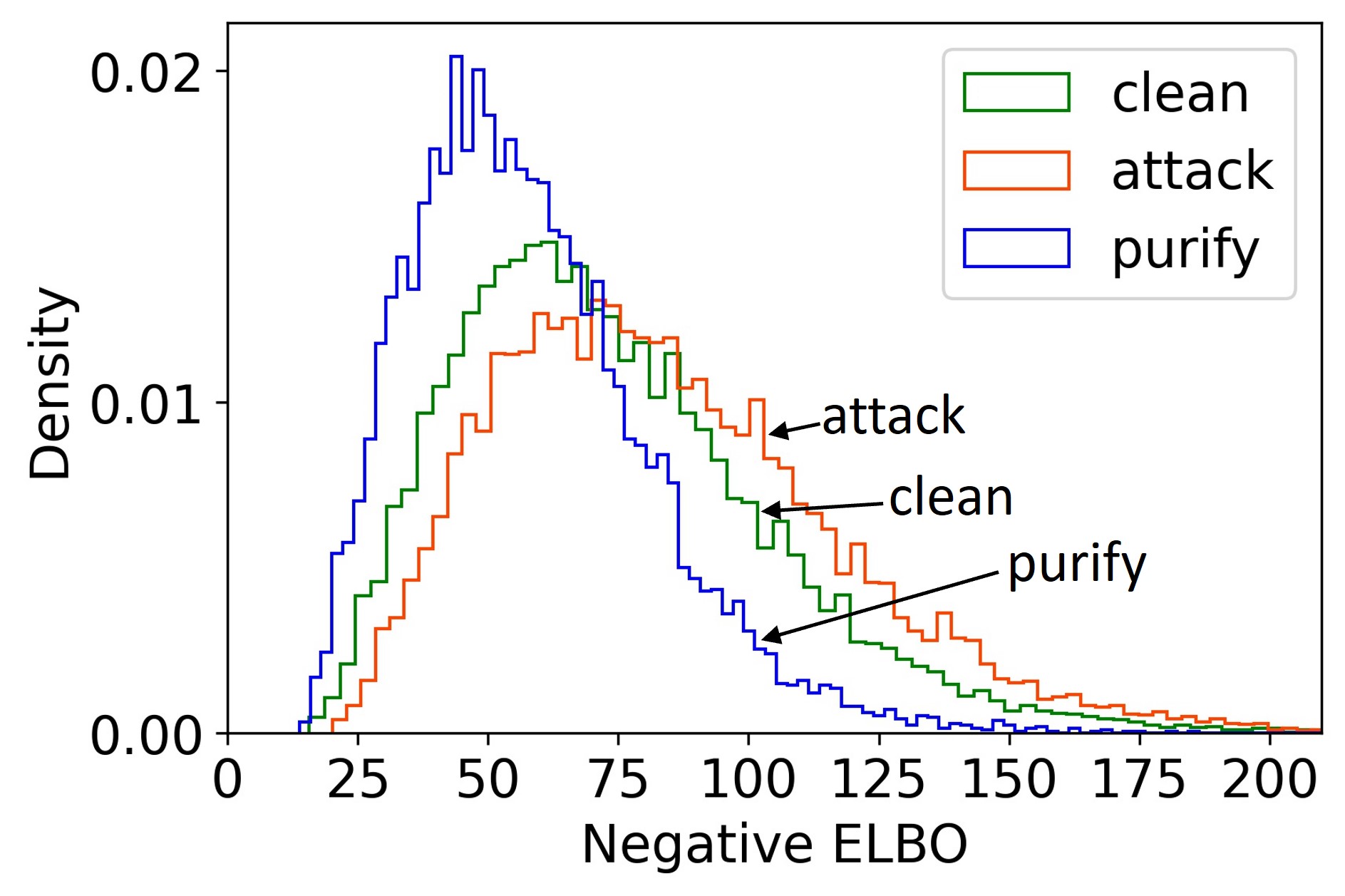}}
\semcaption{Illustration of negative ELBOs.
Adversarial attacks yield higher negative ELBOs while our purification reverses the ELBO shifts. Compared with MNIST, CIFAR-10 has more diverse backgrounds, leading to less obvious ELBO shifts.}
\label{fig:elbo_curve}
\end{figure}

Figure~\ref{fig:cifar_AE_VAE} shows various sample predictions and reconstructions from the VAE-Classifier on CIFAR-10 and CelebA. The results are slightly different from those on MNIST, Fashion-MNIST, and SVHN: in Figures~\ref{fig:cifar_AE_VAE}(a)-(b), although the reconstructed images are more blurry compared with MNIST, Fashion-MNIST, and SVHN, the VAE-Classifier is still robust under adversarial attacks. Distribution of the negative ELBO for clean, attack, and purified examples are shown in Figure~\ref{fig:elbo_curve}.

We use tiny ResNet backbones for ablation study between the Standard-AE-Classifier and the VAE-Classifier. Classification accuracy of CIFAR-10 and SVHN is provided in Table \ref{tbl:cifar_performance} (see Appendix E for results on MNIST and Fashion-MNIST). According to our results, optimizing the ELBO during the test time is more effective than only optimizing the reconstruction loss. Table~\ref{tbl:rev_performacne_rep_network} shows test-time purification (CIFAR-10) with larger backbones such as ResNet-50 (standard training) and PreActResNet-18 (adversarial training). With our defense, the robust accuracy of ResNet-50 on CIFAR-10 increases by more than 50\%. 
Our method can also be applied to adversarially trained models (PreActResNet-18) to further increase their robust accuracy.

\begin{table}[!ht]
\centering
   \setlength{\tabcolsep}{4pt}
	\semcaption{Classification accuracy on SVHN and CIFAR-10 with $\ell_\infty=8/255$ and $\ell_2=0.5$. ``ST-AE'': standard autoencoder, ``CLF'': classifier, ``TTP'': Test-time Purification, ``REC'': minimization of the reconstruction loss, ``ELBO'': minimization of the negative ELBO, and ``AA'': AutoAttack.
 We evaluate the model with both BPDA-(PGD/APGD) and report the minimum accuracy.}
	\resizebox{\linewidth}{!}{
	    \label{tbl:cifar_performance}
		\begin{tabular}{l|l|rrrrrr}
			\hline
			\multicolumn{1}{c|}{Dataset} & \multicolumn{1}{c|}{Method}
			& Clean & FGSM &  PGD & AA-$\ell_\infty$ & AA-$\ell_2$ & BPDA\\
			\hline \hline
			\multirow{5}{*}{SVHN} &ST-AE-CLF&  94.00 & 8.88 &  0.00 & 0.00 & 1.67 & \multicolumn{1}{c}{-}\\
			& +TTP (REC) & 93.03 & 40.10 & 40.83 & 44.82 & 59.93  & 7.65\\
            \cline{2-8} 
			 & VAE-CLF & 95.27 & 70.24 &  16.01 & 0.33 & 6.61 & \multicolumn{1}{c}{-} \\
			& +TTP (REC) &  90.66 & 77.03 & 72.44 &  \textbf{73.92} & 79.15 & \textbf{66.68} \\
			& +TTP (ELBO) & 86.29 & 75.40 & \textbf{72.72} & 73.47 & 76.21 & 64.70 \\
            \hline 
			\multirow{5}{*}{CIFAR-10} & ST-AE-CLF&  90.96 & 6.42 &  0.00 & 0.00 & 0.98 & \multicolumn{1}{c}{-} \\
			& +TTP (REC)  & 87.80 & 19.36 & 11.65 & 13.74 & 44.57 & 0.70 \\
			\cline{2-8}
			 & VAE-CLF & 91.82 & 54.55 &  17.82 & 0.05 & 2.36  & \multicolumn{1}{c}{-}\\
			& +TTP (REC) &  78.51 & 57.24 & 51.20 &  51.63 & 59.35 & 43.02\\
			& +TTP (ELBO) & 77.97 & 59.51 & \textbf{57.21} & \textbf{58.78} & 63.38 & \textbf{47.43}\\
			\hline
		\end{tabular}
	}
\end{table}

\begin{table}[!ht]
\centering
\setlength{\tabcolsep}{1.5pt}
	\semcaption{Classification accuracy on CIFAR-10 using the VAE-Classifier with ResNet-50 and adversarially-trained PreActResNet-18. We set $\ell_\infty=8/255$ and $\ell_2=0.5$.}
	\resizebox{\linewidth}{!}{
	    \label{tbl:rev_performacne_rep_network}
		\begin{tabular}{l|l|rrrrrrr}
			\hline
			\multicolumn{1}{c|}{Backbone} & \multicolumn{1}{c|}{Method} & Clean & FGSM & PGD & AA-$\ell_\infty$ & AA-$\ell_2$ & CW-$\ell_2$ & BPDA\\
			\hline \hline
			\multirow{2}{*}{ResNet-50} & VAE-CLF & 94.82 & 72.10 & 23.84 & 0.04 & 3.42 & 20.21 & \multicolumn{1}{c}{-} \\
			& +TTP (ELBO) & 85.12 & 70.83 & \textbf{63.09} & \textbf{63.16} & 68.21 & 73.63 & \textbf{57.15} \\
            \hline 
			\multirow{2}{*}{\makecell[l]{PreAct-\\ResNet-18}} & VAE-CLF &  87.35 & 66.38 &  61.08 & 58.65 & 65.67 & 66.58 & \multicolumn{1}{c}{-} \\
			& +TTP (ELBO) & 85.14 & 65.74 & \textbf{61.98} & \textbf{63.73} & 70.06 & 73.85 & \textbf{60.52}\\
			\hline
		\end{tabular}
	}
 \end{table}

\begin{table}[!ht]
	\centering
	\semcaption{Benchmark on SVHN and CIFAR-10. Accuracy is directly reported from the respective paper except for the adaptive attack which is reported from \citet{lee2023robust} for diffusion-based purification and \citet{athalye2018obfuscated,croce2022evaluating} for BPDA. Numbers in \tb{bold} are the minimum/robust accuracy.
 ``$*$'': adversarially trained models, ``-'': missing from the references, and ``Adap.A": strongest adaptive attacks. Other works follow different evaluation standards, and we list them in Appendix E. Ours achieves competitive and robust accuracy without adversarial training.}
	\resizebox{\linewidth}{!}{
        \setlength{\tabcolsep}{1pt}
		\begin{tabular}{l|l|cccc}
			\hline
			 & \multicolumn{1}{c|}{Method} & FGSM & PGD & AA-$\ell_\infty$ & AdapA \\
			\hline \hline
      			 \multirow{5}{*}{\rotatebox{90}{SVHN}} & \makecell[l]{Semi-SL*~\cite{mao2021adversarial}}  & - & \tb{62.12} & 65.50 & - \\
			\cline{2-6} 
			& \makecell[l]{PreActResNet-18*~\cite{rice2020overfitting}} & - & \tb{61.00} & - & - \\
			\cline{2-6} 
			& \makecell[l]{Wide-ResNet-28*~\cite{rebuffi2021fixing}} & - & - & \tb{61.09} & - \\
            \cline{2-6} 
			& \makecell[l]{Wide-ResNet-28~\cite{lee2023robust}} & - & - & - & \tb{49.65} \\
			\cline{2-6} 
			& ResNet-Tiny (\textbf{ours}) & 75.40 & 72.72 & 73.47 & \tb{64.70} \\
			\cline{2-6} 
			\hline \hline
		     \multirow{14}{*}{\rotatebox{90}{CIFAR-10}} & \makecell[l]{Wide-ResNet-28~\cite{shi2020online}} & 64.83 & 53.58  & - & \hspace{2mm}\tb{3.70} \\
            \cline{2-6} 
			& \makecell[l]{ResNet~\cite{song2018pixeldefend}} & 46.00 & 46.00 & - & \hspace{2mm}\tb{9.00} \\
            \cline{2-6} 
		   &   \makecell[l]{Wide-ResNet-28~\cite{hill2021}}  & - & 78.91 & - & \tb{54.90} \\
            \cline{2-6} 
            & \makecell[l]{Wide-ResNet-28~\cite{yoon2021}} & - & 85.45 & - & \tb{33.70} \\
            \cline{2-6} 
		  &	\makecell[l]{PreActResNet-18~\cite{mao2021adversarial}} & - & - & \tb{34.40} & - \\
            \cline{2-6} 
            \cline{2-6} 
		  &	\makecell[l]{Semi-SL*~\cite{mao2021adversarial}} & - & 64.44 & 67.79 & \tb{58.40} \\
		 \cline{2-6} 
		 & \makecell[l]{Wide-ResNet-34*~\cite{madry2018towards}} & - & - & \tb{44.04} & - \\
			\cline{2-6} 
		 &	\makecell[l]{Wide-ResNet-34*~\cite{zhang2019theoretically}} & - & - & \tb{53.08} & - \\
			\cline{2-6} 
		 &	\makecell[l]{Wide-ResNet-70*~\cite{wang2023better}} & - & - & \tb{70.69} & - \\
			\cline{2-6} 
		 &	\makecell[l]{Wide-ResNet-70*~\cite{rebuffi2021fixing}} & - & - & \tb{66.56} & - \\
			\cline{2-6} 
		 & \makecell[l]{PreActResNet18*~\cite{rebuffi2021fixing}} & - & - & \tb{58.50} & - \\
			\cline{2-6} 
		 & \makecell[l]{WideResNet-70-16~\cite{nie2022diffusion}} & - & - & 71.29 & \tb{51.13} \\
			\cline{2-6} 
   	 & \makecell[l]{WideResNet-70-16~\cite{lee2023robust}} & - & - & 70.31 & \tb{56.88} \\
			\cline{2-6} 
		 & ResNet-50 (\textbf{ours}) & 70.83 & 63.09 & 63.16 & \tb{57.15} \\
			\hline
		\end{tabular}
	}
    \label{tbl:cifar_benchmark}
\end{table}
 
\begin{table}[!ht]
\centering
\setlength{\tabcolsep}{2.5pt}
	\semcaption{Classification accuracy on CIFAR-100 and CelebA (size: $x^2$) using the VAE-Classifier with $\ell_\infty=8/255$.}
		\resizebox{\linewidth}{!}{
		\label{tbl:celebA}
		\begin{tabular}{l|ccc|ccc}
			\hline
			& \multicolumn{3}{c|}{VAE-CLF} & \multicolumn{3}{c}{+TTD (neg.-ELBO)} \\
			\multicolumn{1}{c|}{Dataset + Backbone} & Clean &  AA-$\ell_\infty$  &  BPDA & Clean &  AA-$\ell_\infty$  & BPDA \\
			\hline \hline
			CIFAR-100 (WResNet-50-2) & 72.37 & 0.10  & \multicolumn{1}{c}{-} \vline & 42.96  & \textbf{26.13} & \textbf{16.87}  \\
               \hline 
			CelebA-64$^2$ (ResNet-50) & 97.86 & 0.28 & \multicolumn{1}{c}{-} \vline & 95.36 & \textbf{90.34} & \textbf{73.77} \\
			\hline
			CelebA-128$^2$ (ResNet-50) & 97.78 & 0.00 & \multicolumn{1}{c}{-} \vline & 96.81 & \textbf{93.91} & \textbf{74.02} \\
			\hline            
		\end{tabular}
	}
    \label{tbl:rev_performacne}
\end{table}

\begin{figure}[!ht]
\centering
\subfloat[Trade-off term]{\label{fig:cifar_tradeoff}\includegraphics[width=0.16\textwidth]{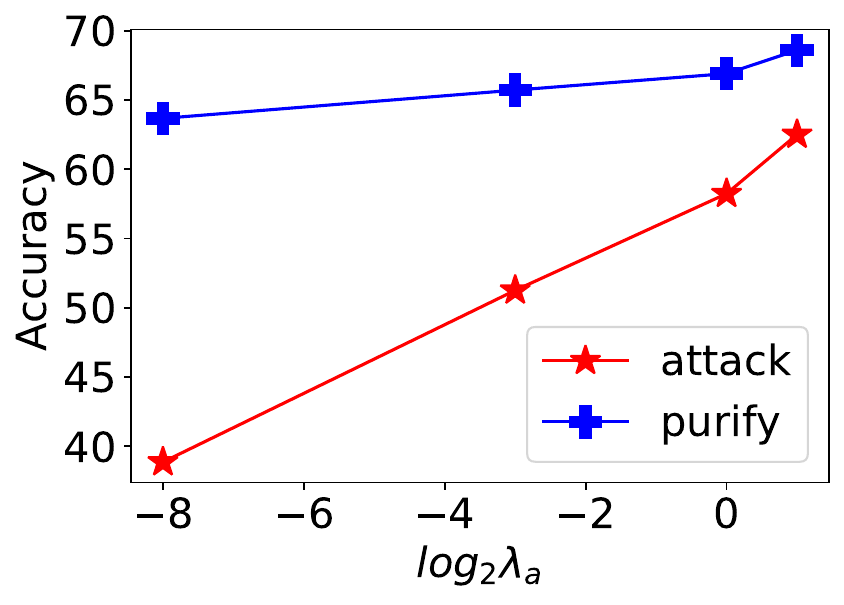}}
\subfloat[Purification budget]{\label{fig:pfy_eps_tradeoff}\includegraphics[width=0.16\textwidth]{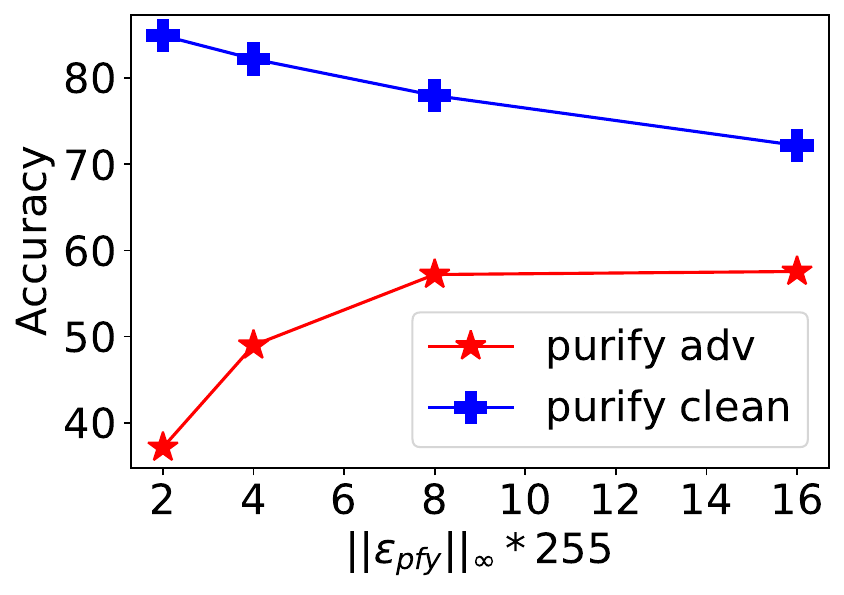}}
\subfloat[Iteration number]{\label{fig:itr_tradeoff}\includegraphics[width=0.16\textwidth]{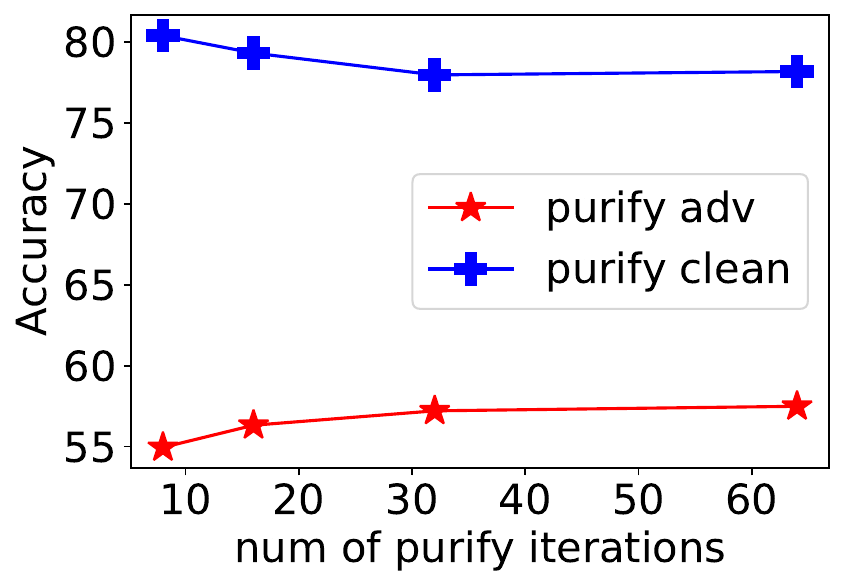}}
\semcaption{Accuracy affected by (a) the trade-off term $\lambda_a$ of the multi-objective attacks, (b) the purification budget $\|\bm \epsilon_{\rm pfy}\|_\infty$ when $\|\bm \delta_{\rm adv}\|_\infty=8/255$, and (c) the number of purification iterations when $\|\bm \delta_{\rm adv}\|_\infty=8/255$ and $\|\bm \delta_{\rm adv}\|_\infty=8/255$.}
\label{fig:ablation_study}
\end{figure}

\noindent\textbf{Multi-objective Attacks.} We evaluate our method with the multi-objective attacks on CIFAR-10 and provide accuracy with respect to trade-off term $\lambda_a$ of Eq.~\eqref{eq:multi_atk_obj_delta_adv} in Figure~\ref{fig:ablation_study}(a). We observe that the classification accuracy of adversarial examples increases as the trade-off term increases while impacts on our defense are not significant. Thus, our defense is robust to the multi-objective attacks. We provide some successful multi-objective adversarial examples in Appendix E. 

\noindent\textbf{Backward Pass Differentiable Approximation (BPDA).} We apply PGD and APGD to optimize the objective of the BPDA attacks. We highlight the minimum classification accuracy from our experiments in Tables~\ref{tbl:cifar_performance}-\ref{tbl:rev_performacne_rep_network}. We observe that although the BPDA attack is the strongest attack compared with the standard adversarial attacks and the multi-objective attacks, our test-time purification still achieves desirable adversarial robustness. In our experiments, models with larger backbones are more robust to the BPDA attacks. Compared with other works in Table \ref{tbl:cifar_benchmark}, our method achieves superior performance against the adaptive white-box attacks.

\noindent\textbf{Blackbox Attacks.}
We evaluate the RayS attack on the ResNet-50 (CIFAR-10) model. For $\ell_\infty$-bounded RayS attack ($\ell_\infty=8/255$), the accuracy increases from 18.02\% to 71.47\% with our defense. For unbounded RayS attack, the accuracy increases from 0.07\% to 70.74\% with our defense. Thus, our defense can provide robustness for blackbox setting as well.

\noindent\textbf{Effects of Hyperparameters.} We study the effects of the purification budgets $\bm \epsilon_{\rm pfy}$ as well as the number of purification iterations on clean classification accuracy and adversarial robustness. We use the VAE-Classifier (with the tiny ResNet backbone) on CIFAR-10 and show results in Figure~\ref{fig:ablation_study} that our defense can provide adversarial robustness with various settings of the hyperparameters.

\noindent\textbf{Larger Datasets.} We use larger datasets to study the impacts of image resolution (CelebA 64$\times$64 and 128$\times$128) and number of classes (CIFAR-100) on our defense. Table~\ref{tbl:celebA} indicates the scalability of
our method for high-resolution data.
However, the performance on data with a larger number of classes is limited as
an accurate estimation of $p(\mathbf{z}|\mathbf{x})$ for each class is required. Compared with 5,000 training images per class in CIFAR-10,
CIFAR-100 only provides 500 per class, leading to a less accurate estimation of $p(\mathbf{z}|\mathbf{x})$. To improve the density estimation of sparse data, one could adapt external data sources such as taxonomy of dataset into the modelling. We defer this to future work. 

\noindent\textbf{Theory and Experiments.} The theory in our methodology section
not only inspires the development of our defense but also provides insights for experimental analysis. For example, CIFAR-10 and SVHN have the same number of classes and dimensions, but our method shows stronger robustness on SVHN. The insight is that reconstruction errors are smaller since the manifold of SVHN is easier to model. Meanwhile, the Standard-AE-Classifier is less robust compared with the VAE-Classifier since the classifier and the decoder of the Standard-AE-Classifier are not semantically consistent. However, the accuracy drops on clean data is not an implication of the theory. There could be multiple reasons for such phenomena. First, optimizing the ELBO for clean images causes distribution drifts (overfitting of the ELBO). Second, there is a tradeoff between robustness and accuracy \cite{tsipras2018robustness}. Third, the function generating the purified signals should be locally Lipschitz continuous with an ideal Lipschitz constant of 1 (Remark~\ref{rk:remark_1}); however, such property is not guaranteed in Eq.~(\ref{eq:iterative_purify}). Consequently, the purification process may move samples to unstable regions causing an accuracy drop. To alleviate this problem, one can apply our method only when attacks are detected.
For instance, when the input reconstructed images and the purified reconstructed images have a large difference, it implies an abnormal reconstruction happens.

\section{Conclusion}
We formulate a novel adversarial purification framework via 
manifold learning and variational inference.
Our test-time purification method is evaluated with several attacks and shows its adversarial robustness for non-adversarially trained models. Even if attackers are aware of our defense method, we can still achieve competitive adversarial robustness. Our method is also capable of defending against adversarial attacks related to VAEs and being combined with adversarially trained models to further increase their adversarial robustness. 

\subsubsection*{Acknowledgments}
Research is supported by DARPA Geometries of Learning (GoL) program under the agreement No. HR00112290075. The views, opinions, and/or findings expressed are those of the authors and should not be interpreted as representing the official views or policies of the Department of Defense or the U.S. government.

\bibliography{aaai24}

\input{supp}

\end{document}

%% file: supp.tex
\setcounter{table}{4}
\setcounter{figure}{7}
\renewcommand{\thefigure}{S.\arabic{figure}}
\renewcommand{\thetable}{S.\arabic{table}}
\renewcommand\thesection{A{}}
\renewcommand\thesubsection{\thesection.\arabic{subsection}}
\clearpage
\appendix
\noindent {\Large{\textbf{Appendix}}}
\section{Section A: Notations}
In this section, we summarize the notations used in the paper. We use bold variables to represent vectors, non-bold variables to represent values. For example, an $n$-dimensional vector can be expressed as $\mathbf{x}=[x_1,...,x_n]^\intercal$. The only \textbf{exception} is the label $\mathbf{y}$. From the theoretical framework section, the letter $\mathbf{y}$ represents integer labels. From the implementation section, the letter $\mathbf{y}$ represents one-hot encoded label vectors. We introduce two element-wise functions for logarithm and sign. We define $\log:\mathbb{R}^n\to\mathbb{R}^n$ as the element-wise logarithm function for vectors, and $\text{sgn}:\mathbb{R}^n\to\mathbb{R}^n$ as the element-wise sign function for vectors. We summarize the key symbols below:

(1) Input image: $\mathbf{x}$, 

(2) Adversarial perturbation: $\bm \delta_{\rm adv}$,

(3) Adversarial example: $\mathbf{x}_{\rm adv}=\mathbf{x}+\bm \delta_{\rm adv}$, 

(4) Purified signal: $\bm \epsilon_{\rm pfy}$,

(5) Purified example: $\mathbf{x}_{\rm pfy}=\mathbf{x}_{\rm adv}+\bm \epsilon_{\rm pfy}$,

(6) Encoder: $\mathbf{f}$,

(7) Decoder: $\mathbf{f}^{\dagger}$,

(8) Classifier on the manifold: $\mathbf{h}$,

(9) Classifier of the image space: $\mathbf{G}$.

\renewcommand\thesection{B{}}
\renewcommand\thesubsection{\thesection.\arabic{subsection}}
\section{Section B: Proofs and Implementations}
Proofs and pseudocode are provided in this section. We briefly recap the concept of the human-level image classifier $\mathbf{G}_R$ and the semantically consistent classifier $\mathbf{h}_S$ on the manifold here. Please refer to our paper for details. 

We define the human-level image classifier $\mathbf{G}_R$ as an approximation of the human vision model. It represents an upper limit of adversarially robust models. Given a clean input $\mathbf{x} \in \mathbb{R}^n$ with a label $\mathbf{y}$, we have $\mathbf{y}=\mathbf{G}_R(\mathbf{x})=\mathbf{G}_R(\mathbf{x}+\bm \eta)$ for every $\|\bm \eta\|_p \leq \tau$ where $\tau$ is considered to be the upper bound of the perturbation budget for static human vision interpretations given an input $\mathbf{x}$.

We define the semantically consistent classifier $\mathbf{h}_S$ on the manifold (with an intrinsic dimension of $m$) as a classifier that is always semantically consistent with the decoder. In other words, for all $\mathbf{z} \in \mathbb{R}^m$, $\mathbf{h}_S(\mathbf{z}) = (\mathbf{G}_R\circ \mathbf{f}^{\dagger})(\mathbf{z})$. 

\subsection{Proofs}
We first introduce the \textbf{triangle inequality of the $\bm \ell_{\bm p}$-metrics}: Given two vectors $\mathbf{a}, \mathbf{b}\in\mathbb{R}^n$, we have $\|\mathbf{a} + \mathbf{b}\|_p \leq \|\mathbf{a}\|_p + \|\mathbf{b}\|_p$. We do not consider $0 < p < 1$ in the following analysis since these p values are uncommon and the inequality may not hold in this range. The $\ell_p$-norm of a vector $\mathbf{a}=[a_1,...,a_n]^\intercal$ is defined as $\|\mathbf{a}\|_p=(\sum_{k=1}^n |a_k|^p)^{1/p}$ for $1 \leq p < \infty$. The $\ell_\infty$-norm is defined as $\|\mathbf{a}\|_\infty=\max\{|a_1|,...,|a_n|\}$ and the $\ell_0$-norm is defined as number of non-zero elements. Based on the Minkowski inequality, we have $\|\mathbf{a} + \mathbf{b}\|_p \leq \|\mathbf{a}\|_p + \|\mathbf{b}\|_p$ for $1 \leq p < \infty$. For the $\ell_\infty$-norm, we have $\|\mathbf{a}+\mathbf{b}\|_\infty = \max\{|a_1+b_1|,...,|a_n+b_n|\} \leq \max\{|a_1|+|b_1|,...,|a_n|+|b_n|\} \leq \|\mathbf{a}\|_\infty + \|\mathbf{b}\|_\infty$. Similarly, we have $\|\mathbf{a} + \mathbf{b}\|_0 \leq \|\mathbf{a}\|_0 + \|\mathbf{b}\|_0$ for the $\ell_0$-norm. Thus, for common p values, we have $\|\mathbf{a} + \mathbf{b}\|_p \leq \|\mathbf{a}\|_p + \|\mathbf{b}\|_p$.

The following proposition states that given an input $\mathbf{x}$, if the human-level image classifier $\mathbf{G}_R$ can be robust up to a perturbation budget of $\tau$ (upper limit for static human vision interpretations), then there exists a function $\mathbf{F}:\mathbb{R}^n \to \mathbb{R}^n$ such that $(\mathbf{h}_S\circ \mathbf{f} \circ \mathbf{F})$ can be robust up to a perturbation budget of $\frac{\tau - \kappa}{2}$ around the input $\mathbf{x}$, where $\kappa$ represents the reconstruction error of $\mathbf{x}$. 

\setcounter{theorem}{0}
\setcounter{proposition}{0}
\setcounter{lemma}{0}
\setcounter{remark}{0}

\begin{proposition}
Let $(\mathbf{x},\mathbf{y})$ be an image-label pair from $\mathcal{D}_{XY}$ and the human-level image classifier $\mathbf{G}_R$ be $(\mathbf{x}, \mathbf{y}, \tau)$-robust. If the encoder $\mathbf{f}$ and the decoder $\mathbf{f}^{\dagger}$ are approximately invertible for the given $\mathbf{x}$ such that the reconstruction error $\|\mathbf{x} - (\mathbf{f}^{\dagger}\circ \mathbf{f})(\mathbf{x})\|_p \triangleq \kappa \leq \tau$ (\textbf{sufficient condition}), then there exists a function $\mathbf{F}:\mathbb{R}^n \to \mathbb{R}^n$ such that $(\mathbf{h}_S\circ \mathbf{f} \circ \mathbf{F})$ is $(\mathbf{x}, \mathbf{y}, \frac{\tau - \kappa}{2})$-robust.
\end{proposition}
The function $\mathbf{F}$ is the defense for attacks. The \textbf{sufficient condition} for $(\mathbf{h}_S\circ \mathbf{f} \circ \mathbf{F})$ to be $(\mathbf{x}, \mathbf{y}, \frac{\tau - \kappa}{2})$-robust is that reconstruction error ($\kappa$) of $\mathbf{x}$ is bounded: $\|\mathbf{x} - (\mathbf{f}^{\dagger}\circ \mathbf{f})(\mathbf{x})\|_p = \kappa \leq \tau$. We assume the sufficient condition hold in the following analysis. To proof the existence of the function $\mathbf{F}$, we first proof the following lemma and theorem.  
\begin{lemma}
\label{lma:lemma_1_app}
If an adversarial example $\mathbf{x}_{\rm adv}=\mathbf{x} + \bm \delta_{\rm adv}$ with $\|\bm \delta_{\rm adv}\|_p \leq  \frac{\tau - \kappa}{2}$ causes $(\mathbf{h}_S \circ \mathbf{f})(\mathbf{x}_{\rm adv}) \neq \mathbf{G}_R(\mathbf{x}_{\rm adv})$, then $\|\mathbf{x}_{\rm adv} - (\mathbf{f}^{\dagger} \circ \mathbf{f})(\mathbf{x}_{\rm adv})\|_p > \frac{\tau+\kappa}{2} \geq \kappa$. 
\end{lemma}
\begin{proof}
For every vector $\bm \xi \in \mathbb{R}^n$ which satisfies $\|\bm \xi\|_p \leq \frac{\tau+\kappa}{2}$, we have $\|(\mathbf{x}_{\rm adv} + \bm \xi) - \mathbf{x}\|_p = \|(\mathbf{x} +\bm \delta_{\rm adv} + \bm \xi)- \mathbf{x}\|_p = \|\bm \delta_{\rm adv} + \bm \xi\|_p \leq \|\bm \delta_{\rm adv}\|_p + \|\bm \xi\|_p \leq \frac{\tau-\kappa}{2}+\frac{\tau+\kappa}{2}=\tau$. It implies that given an adversarial example $\mathbf{x}_{\rm adv}$, for all $\mathbf{\hat{x}} \in \mathbb{R}^n$ and $\|\mathbf{\hat{x}} - \mathbf{x}_{\rm adv}\|_p \leq \frac{\tau+\kappa}{2}$, we have $\mathbf{G}_R(\mathbf{\hat{x}}) = \mathbf{G}_R(\mathbf{x}_{\rm adv}) =\mathbf{G}_R(\mathbf{x})$.

Based on contrapositive of the statement, if $\mathbf{G}_R(\mathbf{\hat{x}}) \neq \mathbf{G}_R(\mathbf{x}_{\rm adv}) =\mathbf{G}_R(\mathbf{x})$, then $\|\mathbf{\hat{x}} - \mathbf{x}_{\rm adv}\|_p > \frac{\tau+\kappa}{2}$. For a semantically consistent classifier $\mathbf{h}_S$, we have $(\mathbf{h}_S \circ \mathbf{f})(\mathbf{x}_{\rm adv}) = (\mathbf{G}_R \circ \mathbf{f}^{\dagger} \circ \mathbf{f})(\mathbf{x}_{\rm adv}) \neq \mathbf{G}_R(\mathbf{x}_{\rm adv})$. Therefore, $\|\mathbf{x}_{\rm adv} - (\mathbf{f}^{\dagger} \circ \mathbf{f})(\mathbf{x}_{\rm adv})\|_p > \frac{\tau+\kappa}{2} \geq \kappa$. 
\end{proof}
The lemma states that successful adversarial attacks on a semantically consistent classifier causes abnormal reconstructions (large reconstruction errors) from the decoder. This criterion could be applied to adversarial attack detection as well. The following theorem states that if a purified sample $\mathbf{x}_{\rm pfy}=\mathbf{x}_{\rm adv} + \bm \epsilon_{\rm pfy}$ has a reconstruction error no larger than $\kappa$, the prediction of $(\mathbf{h}_S \circ \mathbf{f})(\mathbf{x}_{\rm pfy})$ will be the same as the prediction of $\mathbf{G}_R(\mathbf{x})=\mathbf{y}$.
\begin{theorem}
\label{thm:theorem_1_app}
If a purified signal $\bm{\epsilon}_{\rm pfy} \in \mathbb{R}^n$ with $\|\bm{\epsilon}_{\rm pfy}\|_p \leq  \frac{\tau - \kappa}{2}$ ensures that $\|(\mathbf{x}_{\rm adv} + \bm{\epsilon}_{\rm pfy}) - (\mathbf{f}^{\dagger} \circ \mathbf{f})(\mathbf{x}_{\rm adv} + \bm{\epsilon}_{\rm pfy})\|_p \leq \kappa$, then $(\mathbf{h}_S \circ \mathbf{f})(\mathbf{x}_{\rm adv} + \bm{\epsilon}_{\rm pfy}) = \mathbf{G}_R(\mathbf{x})$.
\end{theorem}
\begin{proof}
For every vector $\bm \zeta \in \mathbb{R}^n$ with $\|\bm \zeta\|_p \leq \kappa$, we have $\|(\mathbf{x} +\bm \delta_{\rm adv} + \bm \epsilon_{\rm pfy} + \bm \zeta)- \mathbf{x}\|_p = \|\bm \delta_{\rm adv} + \bm \epsilon_{\rm pfy} + \bm \zeta\|_p \leq \|\bm \delta_{\rm adv}\|_p + \|\bm \epsilon_{\rm pfy}\|_p + \|\bm \zeta\|_p \leq \frac{\tau - \kappa}{2} + \frac{\tau - \kappa}{2} + \kappa = \tau$. It implies that given a purified example $\mathbf{x}_{\rm pfy}=\mathbf{x} +\bm \delta_{\rm adv} + \bm \epsilon_{\rm pfy}$, for all $\mathbf{\hat{x}} \in \mathbb{R}^n$ and $\|\mathbf{\hat{x}} - \mathbf{x}_{\rm pfy}\|_p \leq \kappa$, we have $\mathbf{G}_R(\mathbf{\hat{x}}) = \mathbf{G}_R(\mathbf{x}_{\rm pfy}) = \mathbf{G}_R(\mathbf{x})$.

Thus, $\|(\mathbf{x}_{\rm adv} + \bm{\epsilon}_{\rm pfy}) - (\mathbf{f}^{\dagger} \circ \mathbf{f})(\mathbf{x}_{\rm adv} + \bm{\epsilon}_{\rm pfy})\|_p \leq \kappa$ is a sufficient condition for $\mathbf{G}_R(\mathbf{x}_{\rm adv} + \bm{\epsilon}_{\rm pfy})  = (\mathbf{G}_R \circ \mathbf{f}^{\dagger} \circ \mathbf{f})(\mathbf{x}_{\rm adv} + \bm{\epsilon}_{\rm pfy})$. Based on the definition of $\mathbf{h}_S$, we have $(\mathbf{h}_S \circ \mathbf{f})(\mathbf{x}_{\rm adv} + \bm{\epsilon}_{\rm pfy})=(\mathbf{G}_R \circ \mathbf{f}^{\dagger} \circ \mathbf{f})(\mathbf{x}_{\rm adv} + \bm{\epsilon}_{\rm pfy})$. Thus, $(\mathbf{h}_S \circ \mathbf{f})(\mathbf{x}_{\rm adv} + \bm{\epsilon}_{\rm pfy})=\mathbf{G}_R(\mathbf{x}_{\rm adv} + \bm{\epsilon}_{\rm pfy})=\mathbf{G}_R(\mathbf{x})$.
\end{proof}
Now, we can prove the existence of the defense function $\mathbf{F}$. Let $\mathbf{S}:\mathbb{R}^n \to \mathbb{R}^n$ be a function that takes an input $\mathbf{x}$ and outputs a purified signal $\bm \epsilon_{\rm pfy}=\mathbf{S}(\mathbf{x})$ which minimizes the reconstruction error to a value no larger than $\kappa$, then $\mathbf{F}(\mathbf{x})\triangleq\mathbf{x}+\mathbf{S}(\mathbf{x})$ and $\mathbf{h}_S \circ \mathbf{f} \circ \mathbf{F}$ is $(\mathbf{x}, \mathbf{y}, \frac{\tau - \kappa}{2})$-robust. It is ideal but challenging to achieve $\bm \epsilon_{\rm pfy}=-\bm \delta_{\rm adv}$ since $\bm \delta_{\rm adv}$ is unknown. However, we do expect the function $\mathbf{S}$ to be locally Lipschitz continuous on some neighborhoods. 
\begin{remark}
\label{rk:remark_1_app}
For every perturbation $\bm \delta \in \mathbb{R}^n$ with $\|\bm \delta\|_p \leq \nu$, if $\mathbf{S}(\mathbf{x} + \bm \delta)=-\bm \delta$, then the function $\mathbf{S}:\mathbb{R}^n \to \mathbb{R}^n$ is locally Lipschitz continuous on $\mathcal{B}_\nu \triangleq \{\hat{\mathbf{x}} \in \mathbb{R}^n \mid \|\hat{\mathbf{x}} - \mathbf{x}\|_p < \nu\}$ with a Lipschitz constant of 1.
\end{remark}
\begin{proof}
Given an input $\mathbf{x}$ and two unit vectors $\bm \epsilon_1 \in \mathbb{R}^n$ and $\bm \epsilon_2 \in \mathbb{R}^n$ with $\|\bm \epsilon_1\|_p=\|\bm \epsilon_2\|_p=1$. We define $\mathbf{x}_1 = \mathbf{x} + \alpha_1 \bm \epsilon_1$ and $\mathbf{x}_2 = \mathbf{x} + \alpha_2 \bm \epsilon_2$ where $\alpha_1$ and $\alpha_2$ are values with $0 \leq \alpha_1 \leq \nu$ and $0 \leq \alpha_2 \leq \nu$. Since $\mathbf{S}(\mathbf{x}_1) = -\alpha_1 \bm \epsilon_1$ and $\mathbf{S}(\mathbf{x}_2) = -\alpha_2 \bm \epsilon_2$, then $\|\mathbf{S}(\mathbf{x}_1) - \mathbf{S}(\mathbf{x}_2)\|_p = \|\mathbf{x}_1 - \mathbf{x}_2\|_p$. Therefore, the function $\mathbf{S}$ is locally Lipschitz continuous near the input $\mathbf{x}$ and the Lipschitz constant is 1. 
\end{proof}
Lipschitz continuity is a desire property for the function $\mathbf{S}$, but it is not a necessary condition for adversarial robustness. We do not enforce the Lipschitz continuity in our implementation; however, we conjecture that this leads to the drops on clean classification accuracy. 

\subsection{Algorithm and Implementation}
\label{implementation_limitation}
Finding the function $\mathbf{S}$ is a problem related to functionals and calculus of variations. One could train a model on the domain of adversarial examples to approximate the purified signals, but this approach is computationally expensive. We use the test-time optimization to approximate the function $\mathbf{S}$. The purification objective is expressed as $\mathcal{F}$, which could be the negative reconstruction loss or the ELBO. We apply the element-wise $\min$ and $\max$ functions to project the purified signal into feasible regions. Pseudocode for implemented purification is described in Algorithm \ref{alg:ttd}.

\renewcommand\thesection{C{}}
\renewcommand\thesubsection{\thesection.\arabic{subsection}}
\section{Section C: Datasets}
We perform image classification on 6 datasets: MNIST \cite{lecun2010mnist}, Fashion-MNIST \cite{xiao2017/online}, Street View House Numbers (SVHN) \cite{netzer2011reading}, CIFAR-10 \cite{Krizhevsky09learningmultiple}, CIFAR-100 \cite{Krizhevsky09learningmultiple} and CelebA \cite{liu2015faceattributes}.
The MNIST and Fashion-MNIST datasets contain 60,000 training images and 10,000 testing images.
The CIFAR-10 and CIFAR-100 datasets contain 50,000 training images and 10,000 testing images.
The SVHN dataset is split into 73,257 images for training and 26,032 images for testing.
The CelebA dataset contains 162,770 training images, 19,867 validation images and 19,962 testing images. Each CelebA image has 40 labeled attributes. We evaluate our method on the gender attribute. We downsample the CelebA dataset from a resolution of $178\times218$ to a resolution of $64\times64$ and $128\times128$. See Table~\ref{tb:datasets} for more details.

\begin{table}[htbp]
\centering
\semcaption{Dataset for Image Classification}
\resizebox{\linewidth}{!}{
\label{tb:datasets}
\begin{tabular}{c|r|c|c|r}
\hline
Dataset & Total Number & Resolution & Color & Classes \\ \hline \hline
MNIST & 70,000 & 28$\times$28 & Gray & 10 \\ \hline
Fashion-MNIST & 70,000 & 28$\times$28 & Gray & 10 \\ \hline
SVHN & 99,289 & 32$\times$32 & RGB & 10 \\ \hline
CIFAR-10 & 60,000 & 32$\times$32 & RGB & 10 \\ \hline
CIFAR-100 & 60,000 & 32$\times$32 & RGB & 100 \\ \hline
CelebA-Gender & 202,599 & $64^2 \text{ and } 128^2$ & RGB & 2 \\ \hline
\end{tabular}}
\end{table}

\renewcommand\thesection{D{}}
\renewcommand\thesubsection{\thesection.\arabic{subsection}}
\section{Section D: Architectures and Training Details}

\begin{figure}[!ht]
\centering
\subfloat[Basic Residual Block]{\label{fig:supp_basic_block}\includegraphics[width=0.33\textwidth]{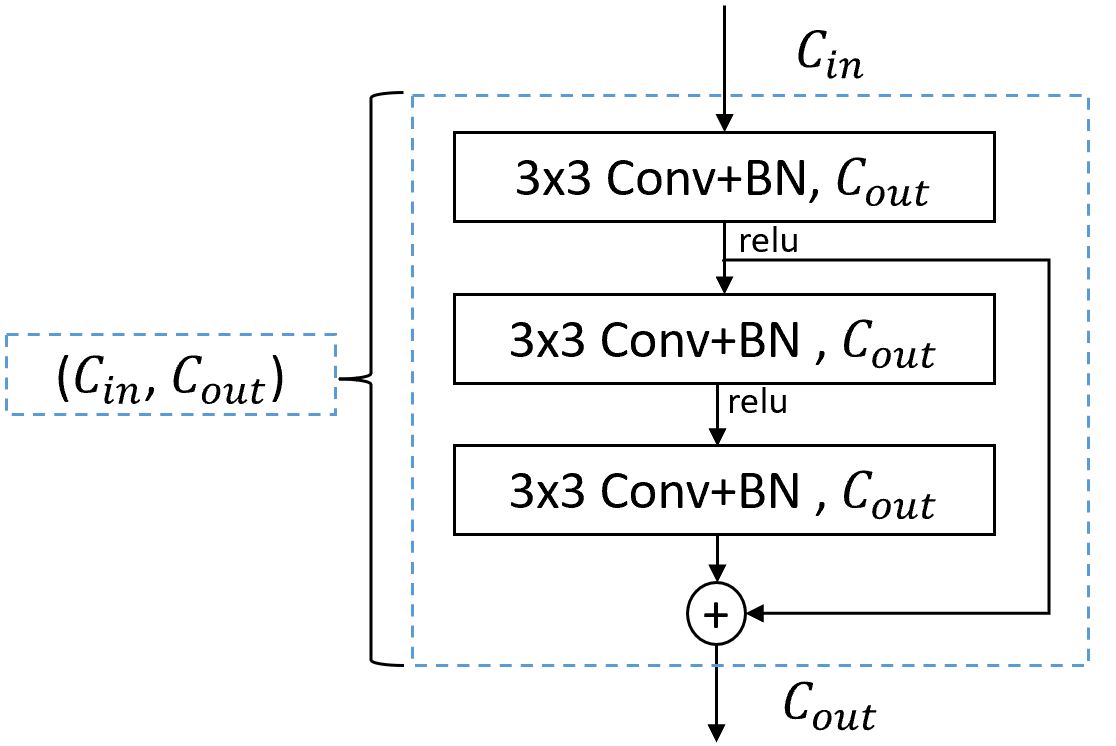}}

\subfloat[Encoder]{\label{fig:encoder}\includegraphics[width=0.23\textwidth]{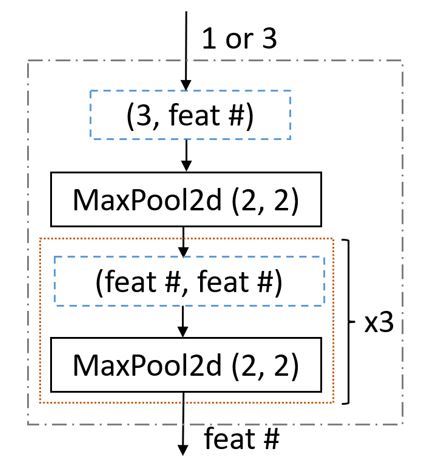}}
\subfloat[Decoder]{\label{fig:decoder}\includegraphics[width=0.256\textwidth]{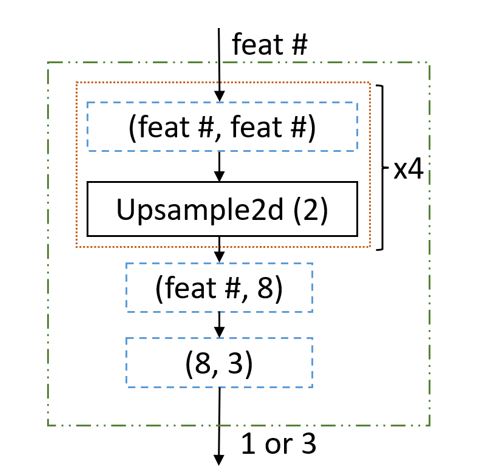}}

\subfloat[Standard-AE-Classifeir]{\label{fig:stae_arch}\includegraphics[width=0.25\textwidth]{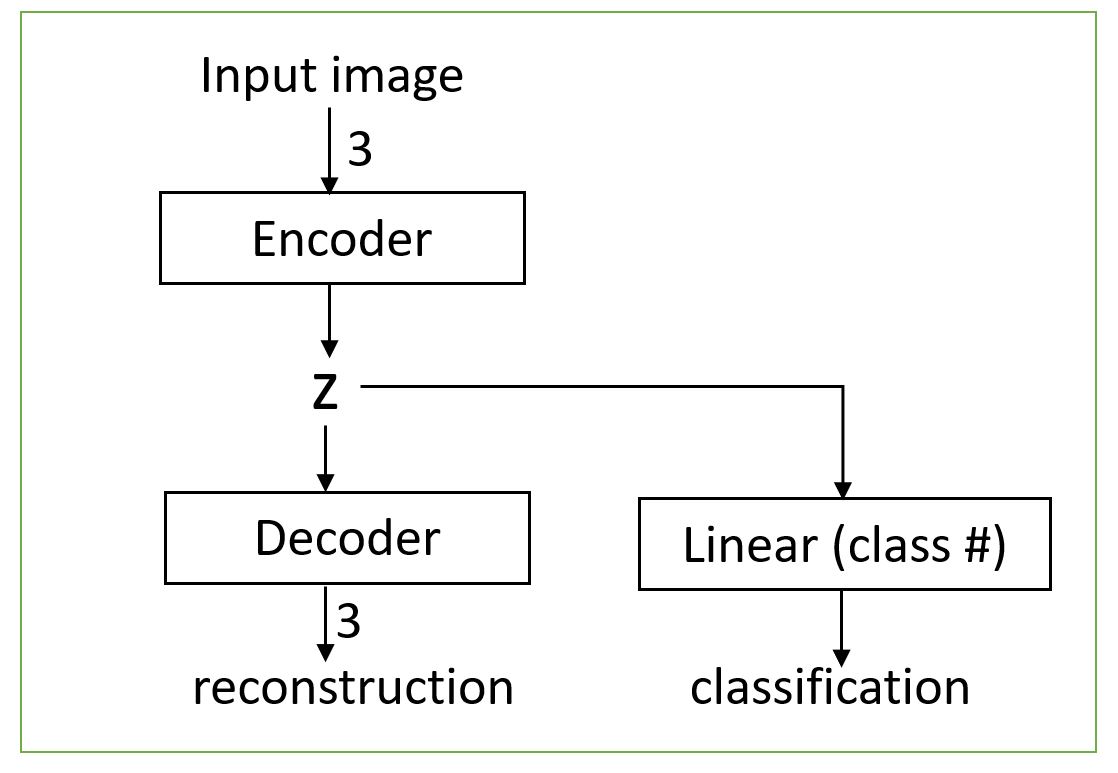}}
\subfloat[VAE-Classifier]{\label{fig:vae_arch}\includegraphics[width=0.206\textwidth]{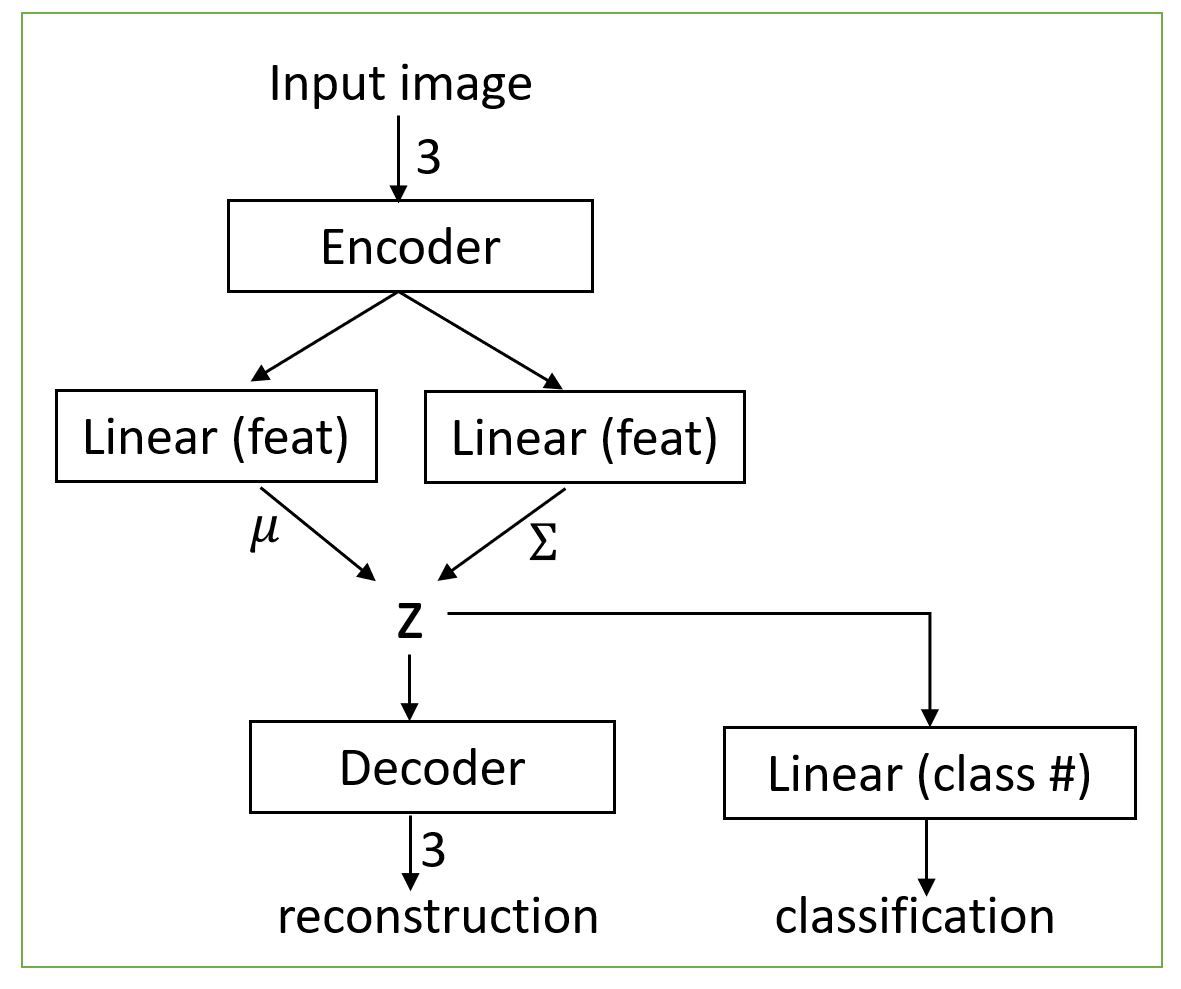}}

\semcaption{Model architectures of the tiny ResNet: (a) basic residual block, (b) encoder, (c) decoder, (d) Standard-AE-Classifier, (e) VAE-Classifier. $C_{\rm in}$ represents number of channels for inputs. $C_{\rm out}$ represents number of channels for outputs. Number of feature maps is expressed as $\text{feat}$ $\#$ in (b) and (c).}
\label{fig:arch}
\end{figure}

For the tiny ResNet (in the ablation study), we use several residual blocks \cite{he2016deep}, see Figure \ref{fig:arch} (a), to construct the encoder and the decoder. The encoder consists of 4 residual blocks, see Figure \ref{fig:arch} (b). The decoder consists of 6 residual blocks, see Figure \ref{fig:arch} (c). We set convolution kernel size to $3 \times 3$ and use max pooling and upsampling to reduce and increase hidden layer dimensions respectively. Number of feature maps (feat \#) varies for each dataset. We use 64 feature maps (in each residual block) for MNIST and Fashion-MNIST, 128 for SVHN, and 256 for CIFAR-10. 

A linear classification head is added on top of the latent vector $\mathbf{z}$ as shown in Figure \ref{fig:arch} (d)-(e). The Standard-AE-Classifier and the VAE-Classifier have similar architectures. The main difference is that the VAE-Classifier samples latent vectors during training. During inference, we use outputs from the mean value ($\mu$) to replace the sampling.

For ResNet-50 (standard training) \cite{he2016deep}, WideResNet-50 (standard training) \cite{DBLP:conf/bmvc/ZagoruykoK16} and PreActResNet-18 (adversarial training) \cite{rebuffi2021fixing}, we construct the decoder with 6 residual blocks (the same as the decoder architecture of the CIFAR-10 tiny ResNet). For the adversarially trained model, we freeze the encoder and only train the decoder.

We use the Adam optimizer \cite{DBLP:journals/corr/KingmaB14} to train models. We set $\beta_1$ to 0.9 and $\beta_2$ to 0.999 for the optimizer. For MNIST and Fashion-MNIST, we train the models for 256 epochs with a learning rate of $10^{-4}$. For SVHN, we train the model for 1024 epochs with a learning rate of $10^{-4}$ and divide it by 10 at the $512^{\text{th}}$ epoch. For CIFAR-10/100, we train the model for 2048 epochs with a learning rate of $10^{-4}$ and divide it by 10 at the $1024^{\text{th}}$ epoch. For CelebA, we train the model for 512 epochs with a learning rate of $10^{-4}$. Batch size for the CelebA model is set to 160 and batch sizes for the rest of the models are set to 256.

\renewcommand\thesection{E{}}
\renewcommand\thesubsection{\thesection.\arabic{subsection}}
\section{Section E: Additional Experiments}
Due to limited space for the manuscript, we present additional experimental results in this section.
\subsection{Performance of MNIST and Fashion-MNIST}
We first show the classification accuracy on MNIST and Fashion-MNIST in Table~\ref{tab:MNIST}. According to the table, test time optimization on ELBO is a more effective defense method for adversarial attacks. Additional visualizations of clean, attack and purified examples of the VAE-Classifier model can be found in Figure~\ref{fig:VAE_atk_app}.
\begin{table}[!ht]
\centering
   \setlength{\tabcolsep}{2pt}
	\semcaption{Classification accuracy on MNIST and Fashion-MNIST. We set attack budget to $\ell_\infty=50/255$ and $\ell_2=3$.}
	\resizebox{\linewidth}{!}{
	    \label{tbl:mnist_performance}
		\begin{tabular}{l|l|rrrrrr}
			\hline
			\multicolumn{1}{c|}{Dataset} & \multicolumn{1}{c|}{Method}
			& Clean & FGSM &  PGD & AA-$\ell_\infty$ & AA-$\ell_2$ & BPDA \\
			\hline \hline
			\multirow{5}{*}{MNIST} & ST-AE-CLF&  99.26 & 44.43 &  0.00 & 0.00 & 0.00 & \multicolumn{1}{c}{-} \\
			& +TTP (REC) & 99.21 & 89.60 & \textbf{90.80} & \textbf{91.21} & 20.80 & 77.41 \\
            \cline{2-8}
			 ~& VAE-CLF & 99.33 & 49.20 &  2.21 & 0.00 & 0.00 & \multicolumn{1}{c}{-} \\
			~& +TTP (REC) &  99.27 & 90.24 & 67.60 &  53.81 & 67.09 & 68.03  \\
			~& +TTP (ELBO) & 99.17 & 93.31 & 86.33 & 85.20 & 81.36  & \textbf{83.08} \\
            \hline 			
			\multirow{4}{*}{F-MNIST} &  ST-AE-CLF &  92.24 & 10.05 &  0.00 & 0.00 & 0.00 & \multicolumn{1}{c}{-}\\
			& +TTP (REC)  & 88.24 & 28.72 & 17.83 & 14.57 & 10.91 & 3.65\\
			\cline{2-8} 
			~ & VAE-CLF & 92.33 & 36.96 &  0.00 & 0.00 & 0.00 & \multicolumn{1}{c}{-} \\
			~& +TTP (REC) &  86.03 & 67.31 & 69.75 &  66.77 & 68.94 & 49.82 \\
			~& +TTP (ELBO) & 84.79 & 71.09 & \textbf{73.44} & \textbf{71.48} & 73.64 & \textbf{60.24} \\
			\hline
		\end{tabular}
	}
 \label{tab:MNIST}
 \end{table}

\begin{figure*}[!ht]
\centering
\setlength{\tabcolsep}{2pt}
\begin{tabular}{m{3mm} m{5.6cm}| m{5.6cm}| m{5.6cm}}
\rotatebox[origin=l]{90}{\footnotesize{Clean}} &
\includegraphics[scale=0.275]{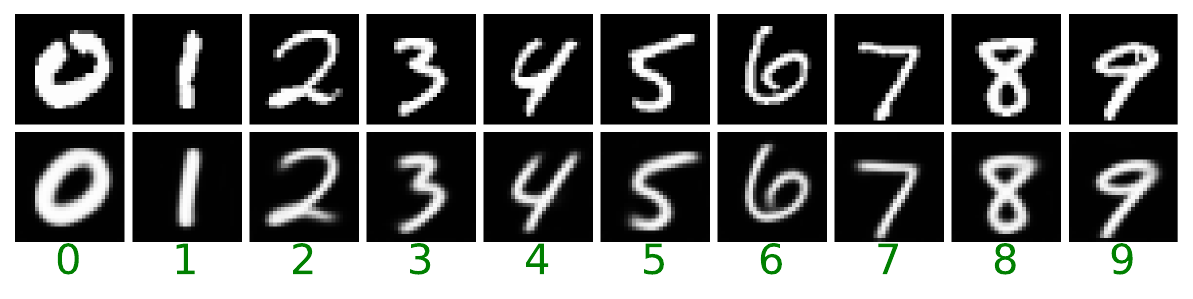} &
\includegraphics[scale=0.275]{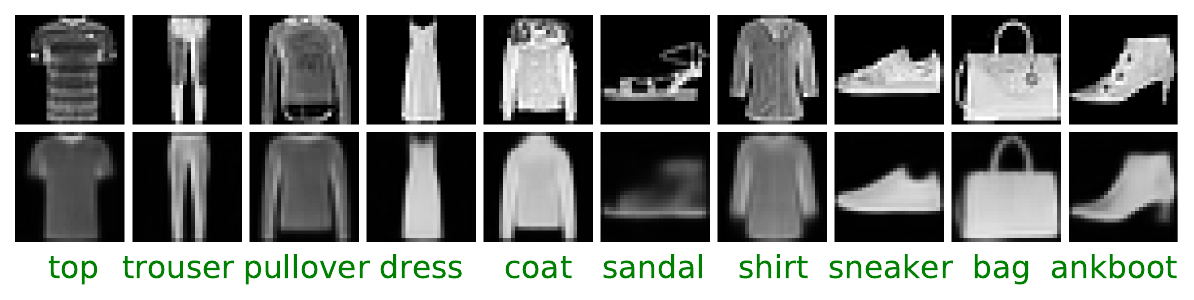} &
\includegraphics[scale=0.275]{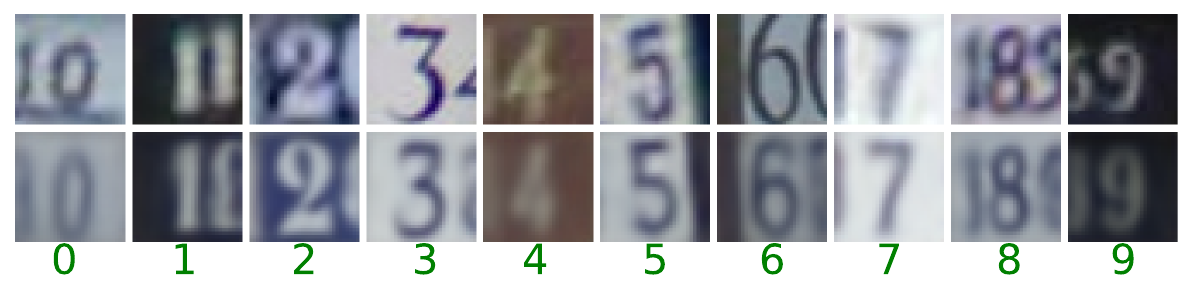} \\
\rotatebox[origin=l]{90}{\footnotesize{Adversarial}} &
\includegraphics[scale=0.275]{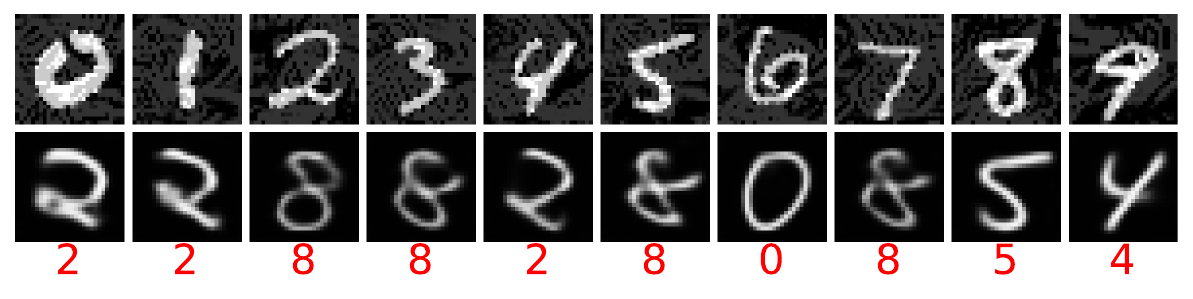} &
\includegraphics[scale=0.275]{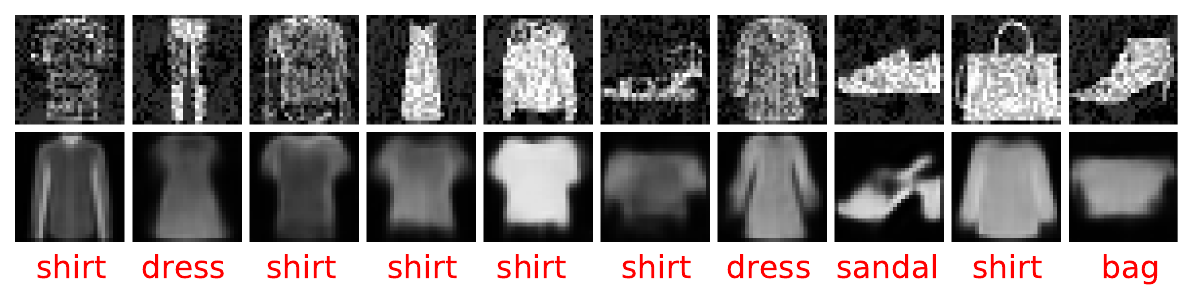} &
\includegraphics[scale=0.275]{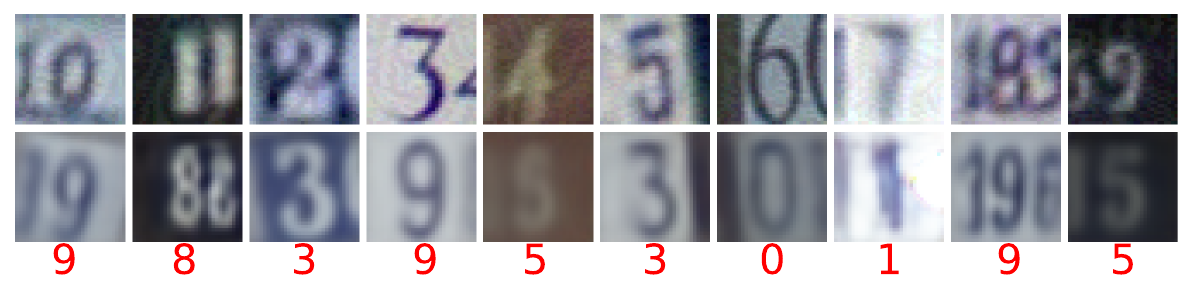} \\
\rotatebox[origin=l]{90}{\footnotesize{Purified}} &
\includegraphics[scale=0.275]{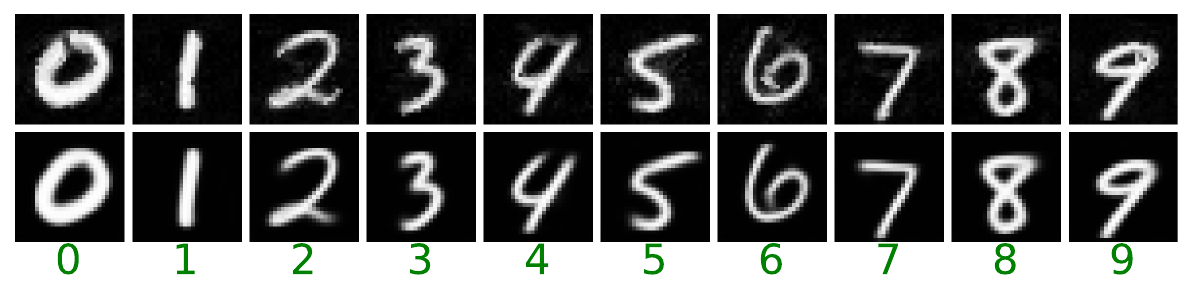} &
\includegraphics[scale=0.275]{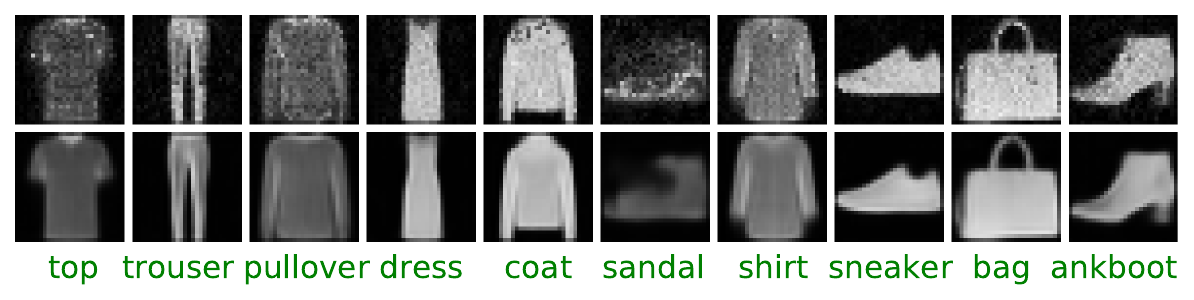} &
\includegraphics[scale=0.275]{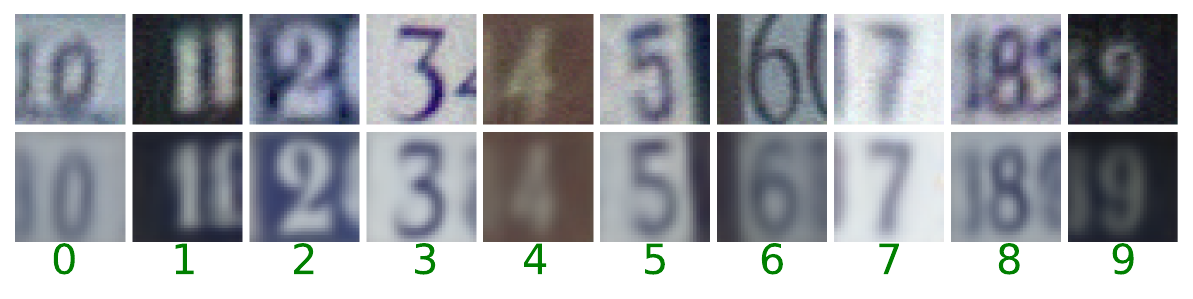} \\
& \multicolumn{1}{c}{\footnotesize{(a) MNIST}} & \multicolumn{1}{c}{\footnotesize{(b) Fashion-MNIST}} & \multicolumn{1}{c}{\footnotesize{(c) SVHN}}
\end{tabular}
\semcaption{Class predictions from the VAE-Classifier models on clean, adversarial and purified samples of (a) MNIST, (b) Fashion-MNIST and (c) SVHN. The top two rows are input and reconstruction of clean images, the middle two rows are input and reconstruction of adversarial images. The bottom two rows are input and reconstruction of purified images. Text represents the predicted classes with {\color{ForestGreen}green} color for correct predictions and {\color{red}red} color for incorrect predictions. Since predictions and reconstructions from the VAE-Classifier are correlated, our test-time purification are effective against adversarial attacks.}
\label{fig:VAE_atk_app}
\end{figure*}

\begin{figure*}[!ht]
\centering
\setlength{\tabcolsep}{2pt}
\begin{tabular}{m{2mm} m{8.5cm}| m{8.5cm}}
\rotatebox[origin=l]{90}{\footnotesize{Clean}} &
\includegraphics[width=0.485\textwidth]{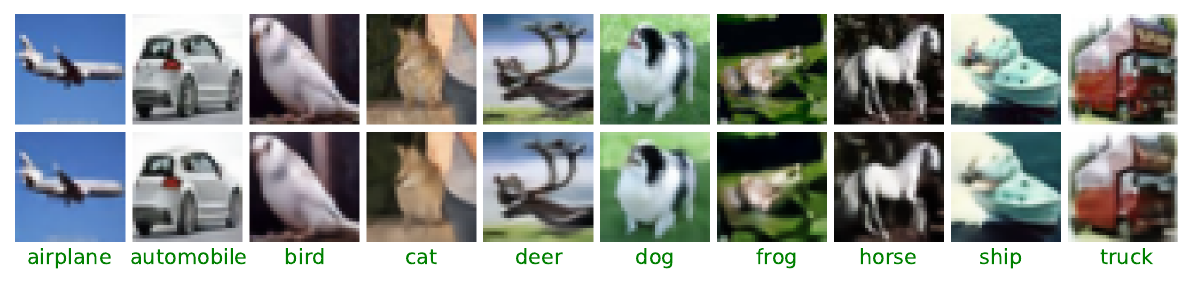} &
\includegraphics[width=0.485\textwidth]{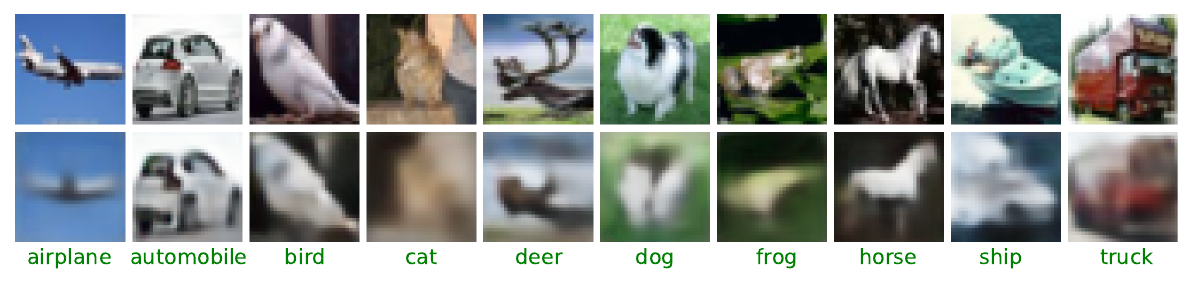} \\
\rotatebox[origin=l]{90}{\footnotesize{Adversarial}} &
\includegraphics[width=0.485\textwidth]{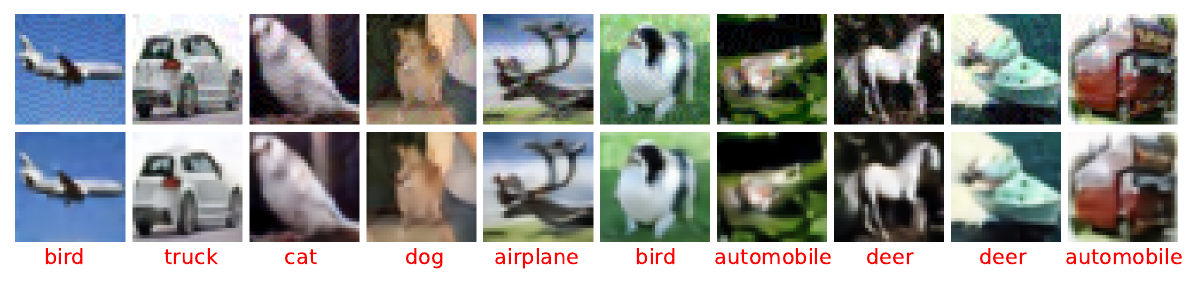} &
\includegraphics[width=0.485\textwidth]{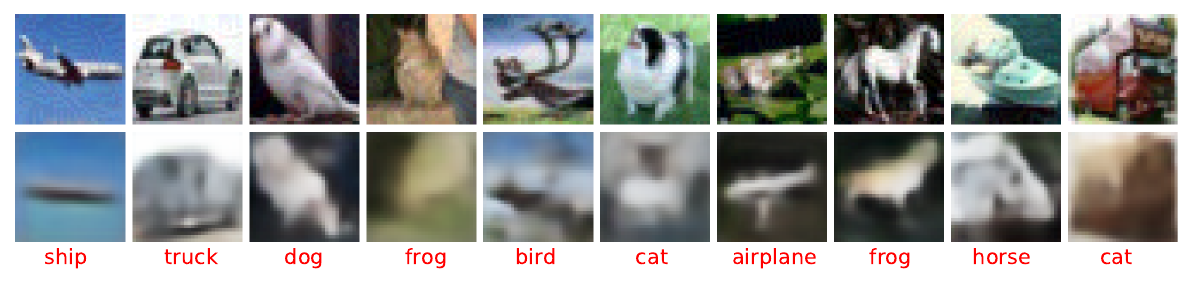} \\
\rotatebox[origin=l]{90}{\footnotesize{Purified}} &
\includegraphics[width=0.485\textwidth]{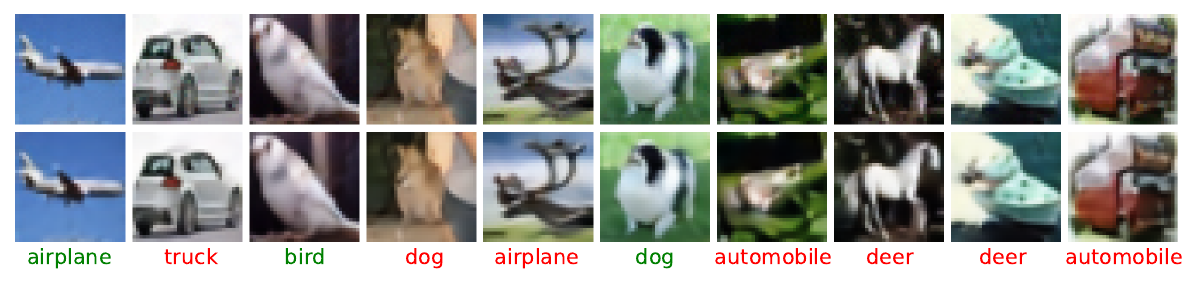} &
\includegraphics[width=0.485\textwidth]{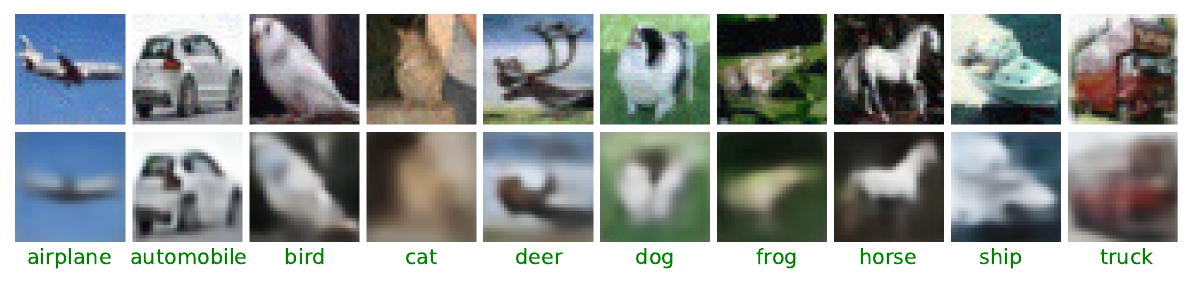} \\
& \multicolumn{1}{c}{\footnotesize{(a) Standard-AE-Classifier}} & \multicolumn{1}{c}{\footnotesize{(b) VAE-Classifier}}
\end{tabular}
\semcaption{Class predictions on clean, adversarial and purified samples of CIFAR-10. The top two rows are input and reconstruction of clean images, the middle two rows are input and reconstruction of adversarial images. The bottom two rows are input and reconstruction of purified images. Text represents the predicted classes with {\color{ForestGreen}green} color for correct predictions and {\color{red}red} color for incorrect predictions. The left column uses the Standard-AE-Classifier and the right column uses the VAE-Classifier. Compared with the VAE-Classifier, the Standard-AE-Classifier generates better reconstructed images; however, there is no semantic consistency (during attacks) between its classifier and its decoder. Therefore, our defense is less effective on the Standard-AE-Classifier.}
\label{fig:cifar_AE_VAE_app}
\end{figure*}

\begin{figure*}[!ht]
\centering
\subfloat[MNIST Recon. Loss]{\label{fig:mnist_rec}\includegraphics[width=0.24\textwidth]{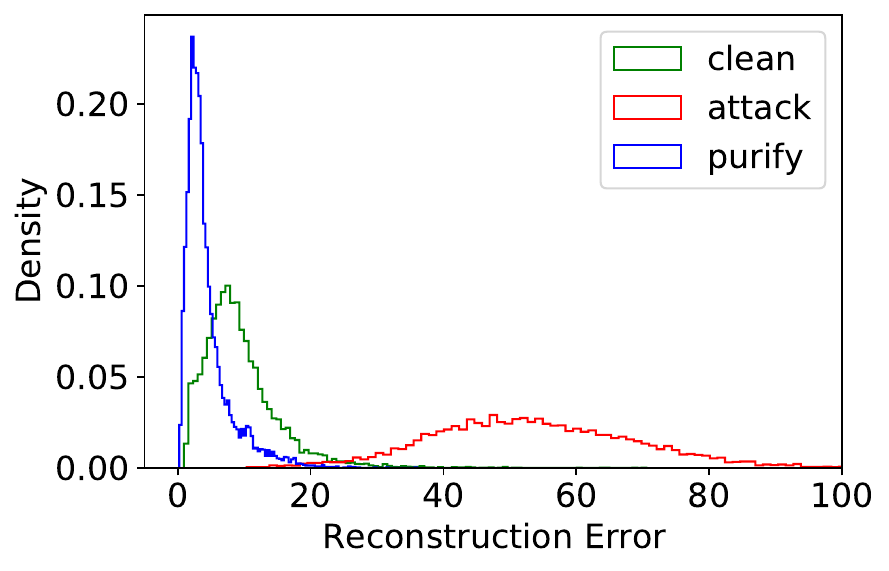}}
\subfloat[MNIST KL Divergence]{\label{fig:mnist_kl}\includegraphics[width=0.24\textwidth]{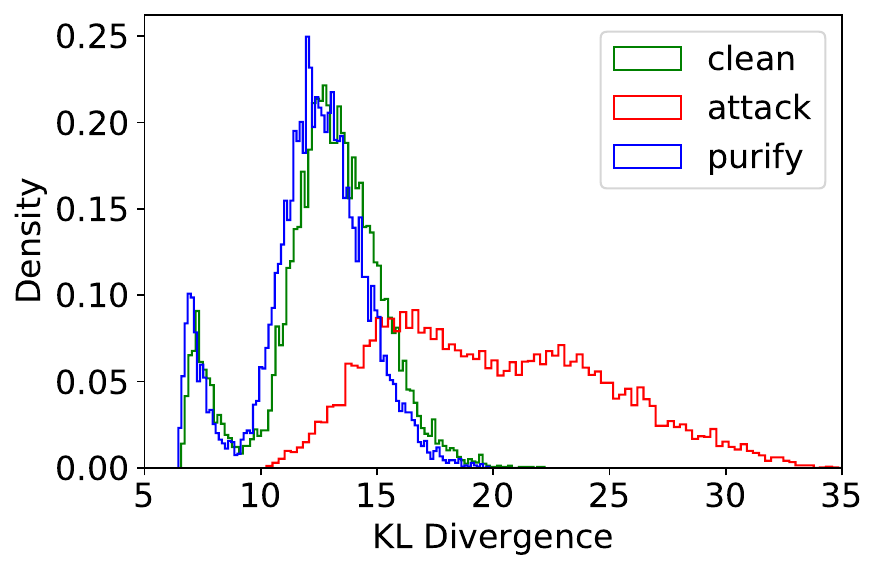}}
\subfloat[F-MNIST Recon. Loss]{\label{fig:fmnist_rec}\includegraphics[width=0.24\textwidth]{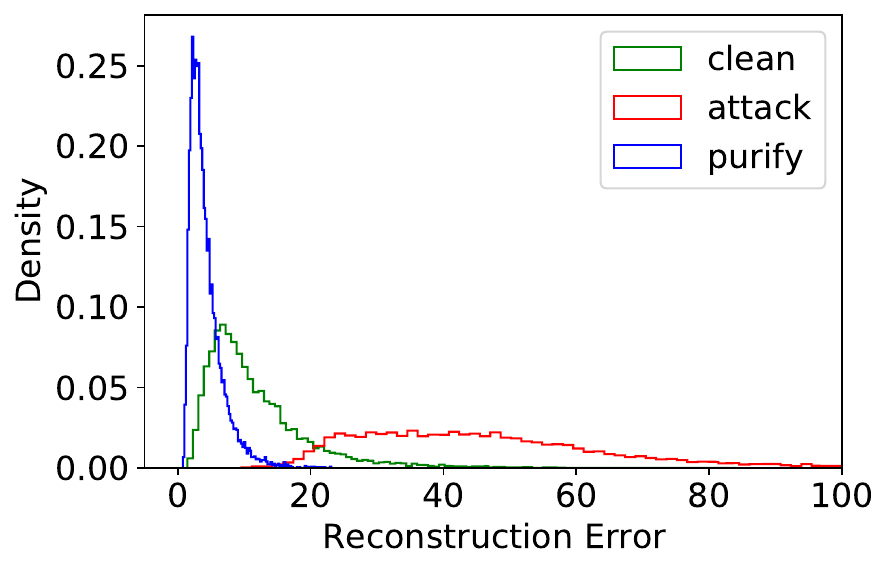}}
\subfloat[F-MNIST KL Divergence]{\label{fig:fmnist_kl}\includegraphics[width=0.24\textwidth]{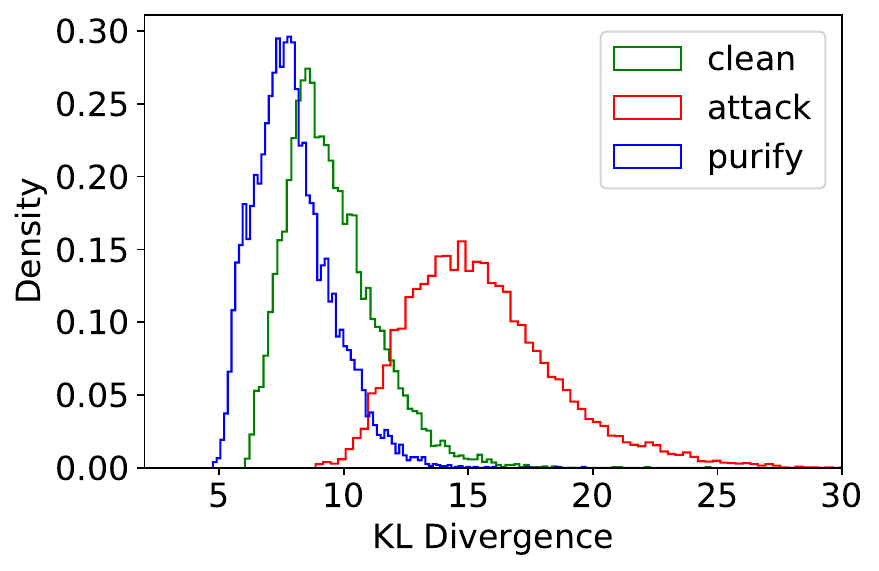}}

\subfloat[SVHN Recon. Loss]{\label{fig:svhn_rec}\includegraphics[width=0.24\textwidth]{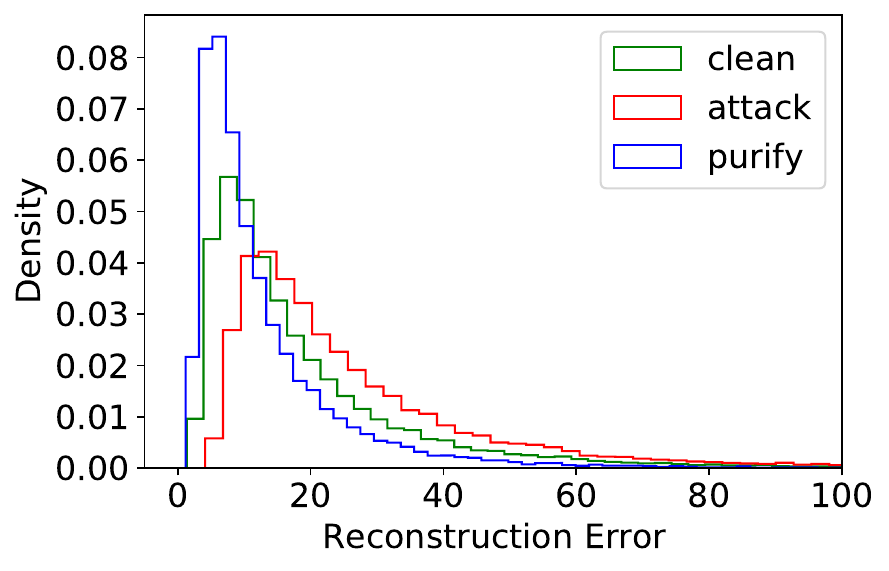}}
\subfloat[SVHN KL Divergence]{\label{fig:svhn_kl}\includegraphics[width=0.24\textwidth]{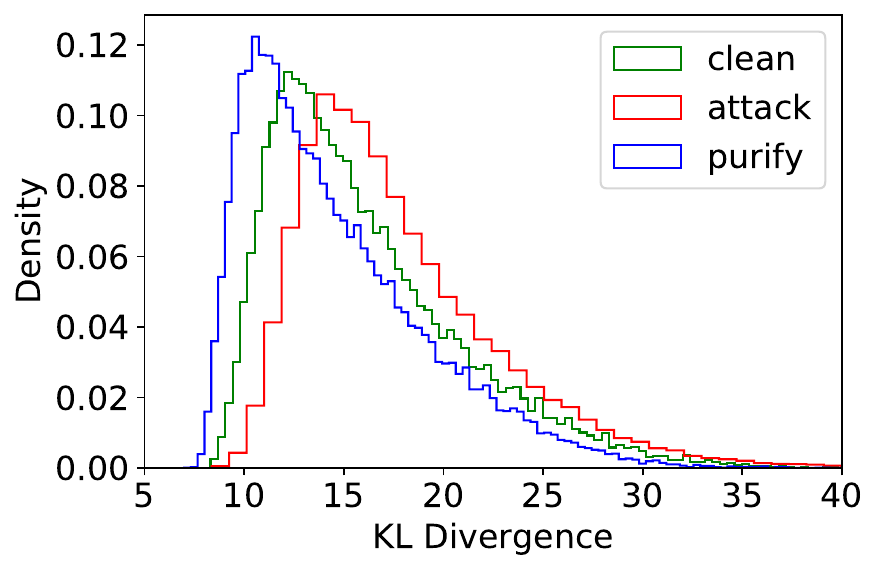}}
\subfloat[CIFAR-10 Recon. Loss]{\label{fig:cifar_rec}\includegraphics[width=0.24\textwidth]{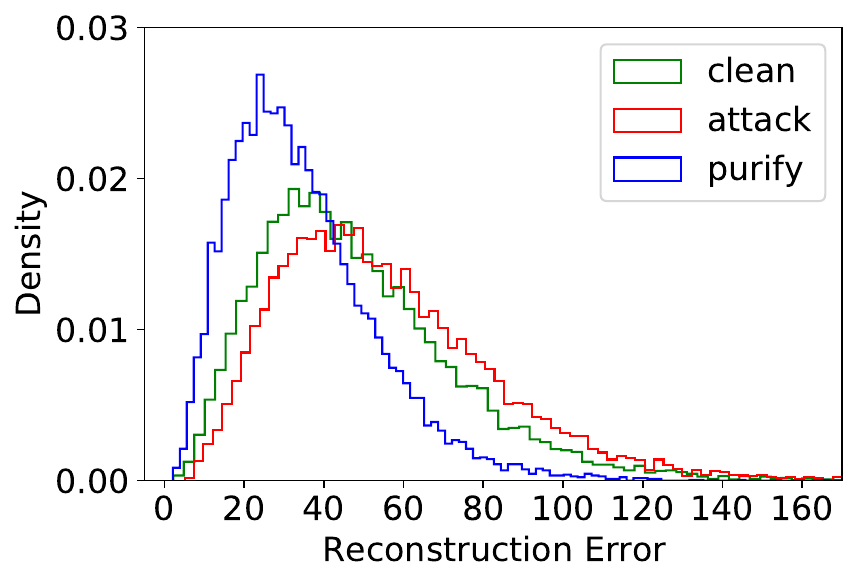}}
\subfloat[CIFAR-10 KL Divergence]{\label{fig:cifar_kl}\includegraphics[width=0.24\textwidth]{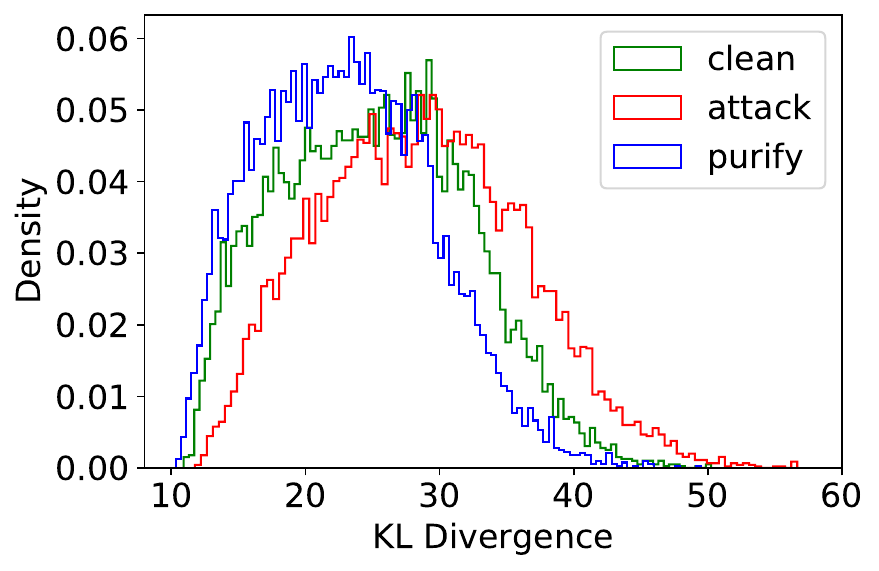}}
\semcaption{Distributions of the reconstruction loss and the KL divergence on clean, adversarial and purified examples.}
\label{fig:rec_KL_curve}
\end{figure*}

\begin{figure}[!ht]
\centering
\setlength{\tabcolsep}{2pt}
\begin{tabular}{m{4cm} m{4cm}}
    \includegraphics[width=0.2\textwidth]{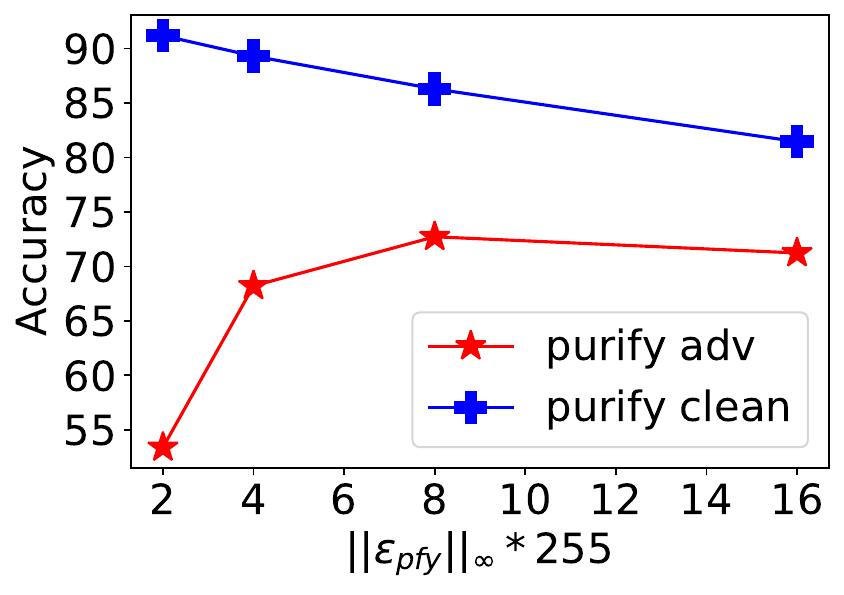}
    &\includegraphics[width=0.2\textwidth]{fig/cifar_eps.pdf}\\
    \includegraphics[width=0.2\textwidth]{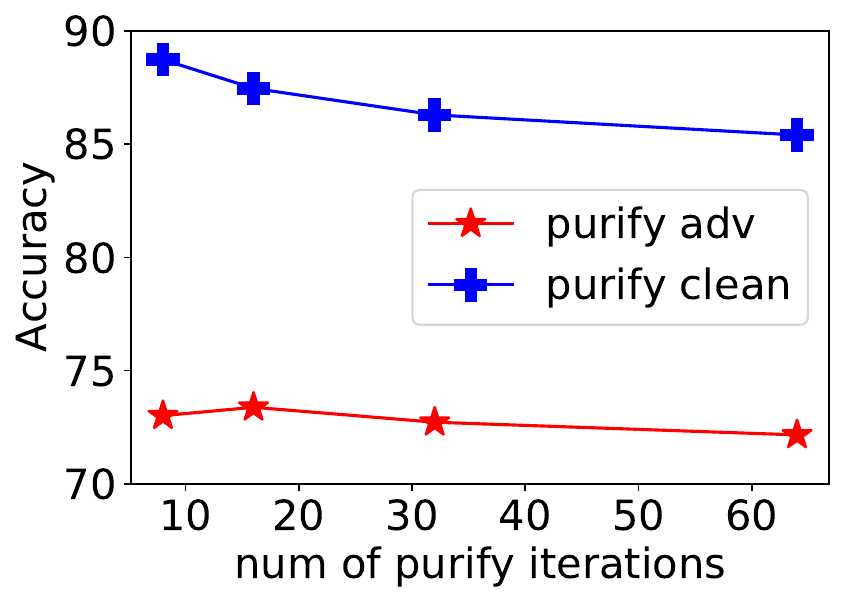}
    &\includegraphics[width=0.2\textwidth]{fig/cifar_itr.pdf}\\
    \includegraphics[width=0.2\textwidth]{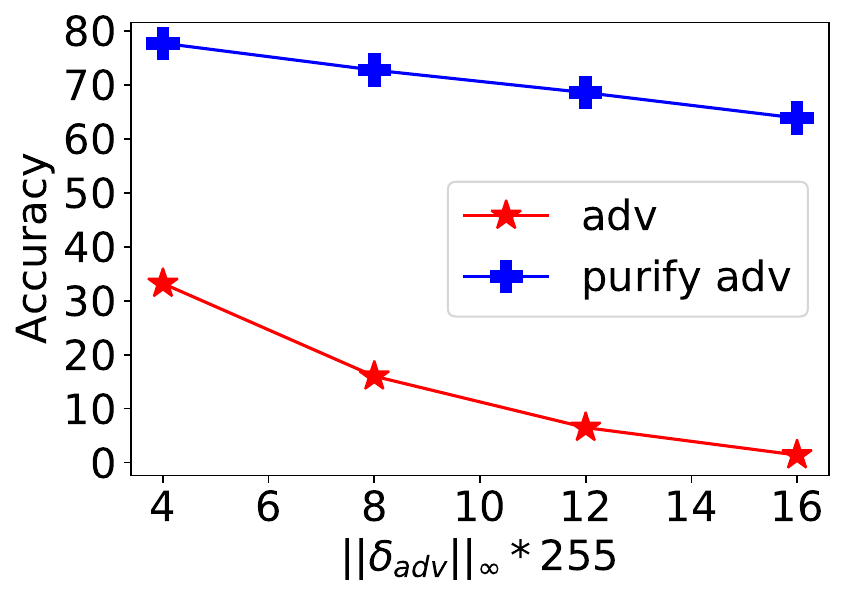}
    &\includegraphics[width=0.2\textwidth]{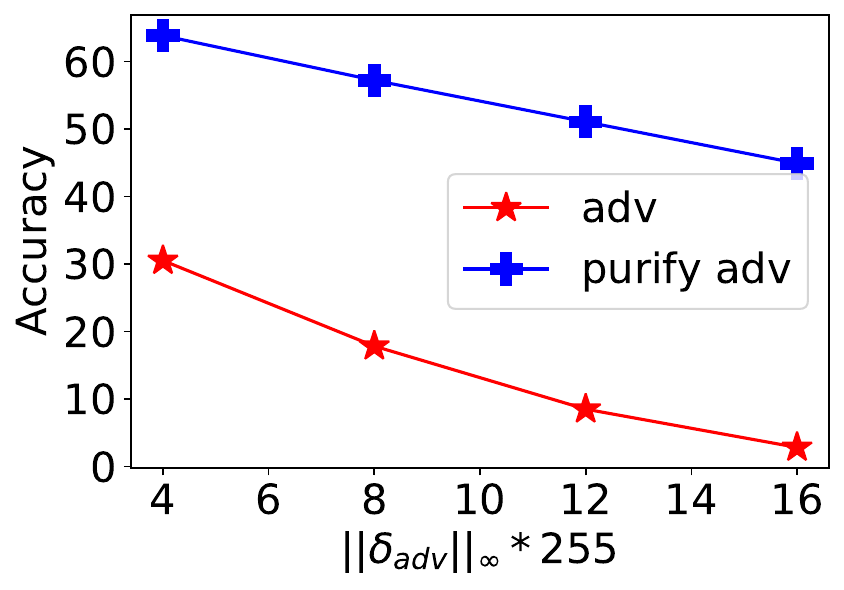}\\
    \multicolumn{1}{c}{\footnotesize{(a) SVHN}} & \multicolumn{1}{c}{\footnotesize{(b) CIFAR-10}}
    \end{tabular}
\semcaption{Effects of hyperparameters on classification. \textbf{Top}: Given adversarial perturbations of $\ell_\infty=8/255$, the change of accuracy with respect to the $\ell_\infty$ purification budgets. \textbf{Middle}: Given adversarial perturbations of $\ell_\infty=8/255$ and a purification budget of $\ell_\infty=8/255$, the change of accuracy with respect to the number of purification iterations. \textbf{Bottom}: Given a purification budget of $\ell_\infty=8/255$ and 32 purification iterations, the change of accuracy with respect to $\ell_\infty$ adversarial perturbation budgets.}
\label{fig:ablation_study_app}
\end{figure}

\begin{figure}[!ht]
\centering
\includegraphics[width=0.35\textwidth]{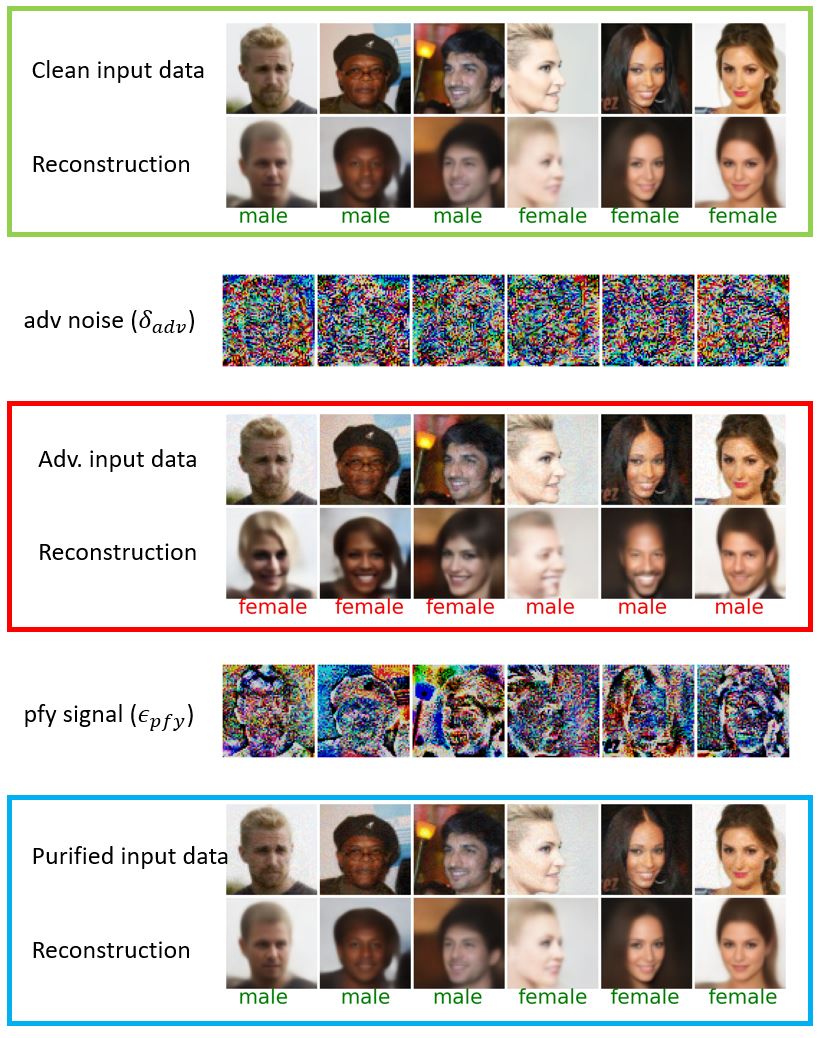}
\semcaption{Examples of adversarial noises and purified signals from the CelebA dataset. Both adversarial noises and purified signals are multiplied by 25 for better visualization. We scale the pixel value to a range of [0, 255] and calculate the mean squared error (MSE). The MSE between the clean reconstructed images and the adversarial reconstructed images is \textbf{416.72}. The MSE between the clean and the purified reconstructed images is \textbf{33.56}.}
\label{fig:celebA_pfy}
\end{figure}

\begin{figure}[!ht]
	\begin{center}
\includegraphics[width=0.65\linewidth]{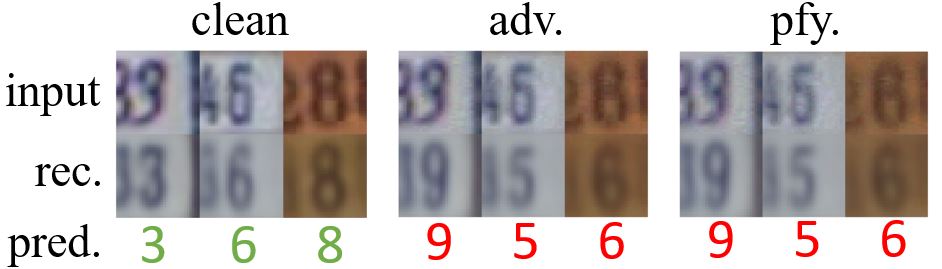}
	\end{center}
	\semcaption{Successful examples of multi-objective attacks.}
\label{fig:vis_multi_obj}
\end{figure}


\subsection{Additional Benchmark}
In this work, we focus on providing adversarial robustness without the need for adversarial training. In table 3 of the main paper, we compare our method with different adversarial training and purification works following the similar evaluation standard. Numbers of the table 3 in the main paper are reported from the receptive paper except the adaptive attack. Adaptive attack from \citet{nie2022diffusion} are reported from \citet{lee2023robust}. Adaptive attack from \citet{song2018pixeldefend} are reported from \citet{athalye2018obfuscated} and adaptive attack from \citet{shi2020online}, \citet{mao2021adversarial} as well as \citet{yoon2021} are reported from \citet{croce2022evaluating}. Next, we provide some additional benchmark which are not listed in the table 3.

Noting that prior purification works (with adversarially-trained models) also evaluate non-adversarially trained models on CIFAR10: \citet{mao2021adversarial} reports 34.40\% auto-attack ($\ell_\infty=8/255$) accuracy on ResNet18, \citet{wang2021fighting} reports 45.4\% auto-attack ($\ell_\infty=1.5/255$) accuracy on ResNet26, \citet{DBLP:conf/iccvw/PerezAJRTGA21} reports 29.81\% ($\ell_\infty=2/255$) robust accuracy on ResNet18. Since \citet{DBLP:conf/aaai/AlfarraPTBTG22} relies on pseudo-label, in the worst case, the robust accuracy of applying their defense on our ResNet50 model is 0.04\%. Our robust accuracy ($\ell_\infty=8/255$) is 47.43\% on tiny ResNet (smaller than ResNet18) and 57.15\% on ResNet-50. For $\ell_\infty=16/255$, robust accuracy (PGD) of tiny ResNet (CIFAR-10) is $44.91\%$ which is higher than $35.49\%$ from PuVAE \cite{hwang2019}. Thus, our approach can achieve higher robustness on smaller models.

\subsection{Standard-AE-Classifier and VAE-Classifier}
Figure \ref{fig:cifar_AE_VAE_app} shows various predictions and reconstructions on clean, adversarial and purified examples using the Standard-AE-Classifier and the VAE-Classifier. For the VAE-Classifier, the abnormal reconstructions during attacks are correlated with the abnormal predictions from the classifier, which is not applied to the standard-AE-Classifier. Thus, our defense is more effective on the VAE-Classifier.

\subsection{Reconstruction Loss and KL Divergence}

The ELBO consists of the reconstruction loss and the KL divergence. We observe that adversarial attacks increase both the reconstruction loss and the KL divergence. It implies that adversarial attacks could create abnormal reconstructions (large reconstruction loss) as well as push latent vectors to low density regions (large KL divergence). We visualize the distributions in Figure \ref{fig:rec_KL_curve}. 

\subsection{More Results for Effects of Hyperparameters}
We provide more results for effects of hyperparameters in Figure~\ref{fig:ablation_study_app}. We use the VAE-Classifier (with the tiny ResNet backbone). According to the figure, our defense is robust over different hyperparameters.

\subsection{Adversarial Noises and Purified Signals}
We provide some examples of adversarial noises and purified signals in Figure~\ref{fig:celebA_pfy}. Adversarial noises and purified signals are somehow correlated but not exactly the same.

We scale the pixel values to a range of $[0,255]$ and calculate the mean squared error (MSE) between the reconstructed images. The MSE between the clean reconstructed images and the purified reconstructed images is 33.56 while the MSE between the clean reconstructed images and the adversarial reconstructed images is 416.72. This phenomenon implies that the purification process indeed move the purified latent vectors closer to the clean latent vectors since the reconstructions from the clean images and the reconstruction from the purified images are similar.

\subsection{Visualization of Multi-objective Adversarial Examples}

The multi-objective attack is less effective than other white-box attacks, but the generated adversarial examples are interesting (see Figure \ref{fig:vis_multi_obj} for examples). Compared with other attacks, the perturbations of multi-objective attacks seem to be more related to semantics of the images (similar to on-manifold adversarial examples).